\pdfoutput=1

\documentclass[11pt]{article}

\usepackage[final]{acl}

\usepackage{times}
\usepackage{latexsym}

\usepackage{booktabs}
\usepackage{multirow}
\usepackage{colortbl}
\usepackage{makecell}

\NewExpandableDocumentCommand\Gap{}{\XGap{3pt}}

\usepackage[T1]{fontenc}

\usepackage[utf8]{inputenc}

\usepackage{microtype}

\usepackage{inconsolata}

\usepackage{graphicx}
\usepackage{amssymb} 
\usepackage{caption} 
\usepackage{amsmath}
\usepackage{amsthm}
\usepackage{bbm}
\usepackage{subcaption}

\DeclareMathOperator*{\minimize}{minimize}

\newtheorem*{definition*}{Definition}
\newtheorem{lemma}{Lemma}
\newtheorem{sublemma}{Sublemma}[lemma] 

\theoremstyle{definition}

\theoremstyle{remark}

\usepackage{nicefrac,xfrac}

\title{Rotate, Clip, and Partition: Towards W2A4KV4 Quantization by Integrating Rotation and Learnable Non-uniform Quantizer}

\author{Euntae Choi\thanks{indicates equal contribution}, Sumin Song\footnotemark[1], Woosang Lim, Sungjoo Yoo\thanks{indicates corresponding author} \\
  Seoul National University \\
  \texttt{euntae.choi175@gmail.com, songsm921@snu.ac.kr,} \\ \texttt{ftyg656512@snu.ac.kr, sungjoo.yoo@gmail.com} \\}

\begin{document}
\maketitle
\begin{abstract}
We propose Rotate, Clip, and Partition (RCP), a Quantization-Aware Training (QAT) approach that first realizes extreme compression of LLMs with W2A4KV4 (2-bit weight, 4-bit activation, and 4-bit KV-cache) configuration. RCP integrates recent rotation techniques with a novel non-uniform weight quantizer design by theoretically and empirically analyzing the impact of rotation on the non-uniformity of weight distribution. Our weight quantizer, Learnable Direct Partitioning (LDP), introduces learnable parameters to directly learn non-uniform intervals jointly with LLM weights. We also present a GPU kernel supporting GEMV on non-uniform W2A4 as proof of concept. Experiments show that RCP can compress LLaMA-2-7B to W2A4KV4 with a loss of only 2.84 WikiText2 PPL and 5.29 times reduced memory footprint. Furthermore, RCP can quantize challenging mobile-targeted LLaMA-3.2 models and domain-specific WizardCoder-7B and MetaMath-7B with no critical problems such as convergence failure and repetition. Code is available at \url{https://github.com/songsm921/RCP}.
\end{abstract}

\section{Introduction}

\label{sec:intro}
Large language models (LLMs) have made significant advancements, but their growing size and resource demands create challenges for deployment across data centers and mobile devices. To address these constraints, extensive research efforts have focused on improving quantization algorithms.

Notably, rotation-based Post-Training Quantization (PTQ) \cite{quarot,spinquant,duquant} showed remarkable improvement on W4A4KV4\footnote{We call $l$-bit weight, $m$-bit activation, and $n$-bit KV-cache W$l$A$m$KV$n$ like W2A4KV4.} quantization, and state-of-the-art Quantization-Aware Training (QAT) \cite{bitdistiller,efficientqat} made extreme weight quantization possible via careful design of datasets and training procedures.

In this work, we propose Rotate, Clip, and Partition (RCP), a rotation-based QAT algorithm, to push the boundaries of extremely low-bit compression. Based on empirical and theoretical analysis, we draw our main insight that rotating LLM weights has two effects at once: eliminating outliers and increasing the non-uniformity of the weight distribution. The key component of RCP is Learnable Direct Partitioning (LDP), which is a fully differentiable non-uniform weight quantizer working in three steps: 1) quantization range setup with learnable clipping parameters \cite{omniquant}; 2) non-uniform quantization via splitting the quantization range with learnable partitioning parameters; 3) non-uniform dequantization that maps the quantized weights to real-valued grids.
\begin{figure}[t!] 
    \centering
    \includegraphics[width=\columnwidth]{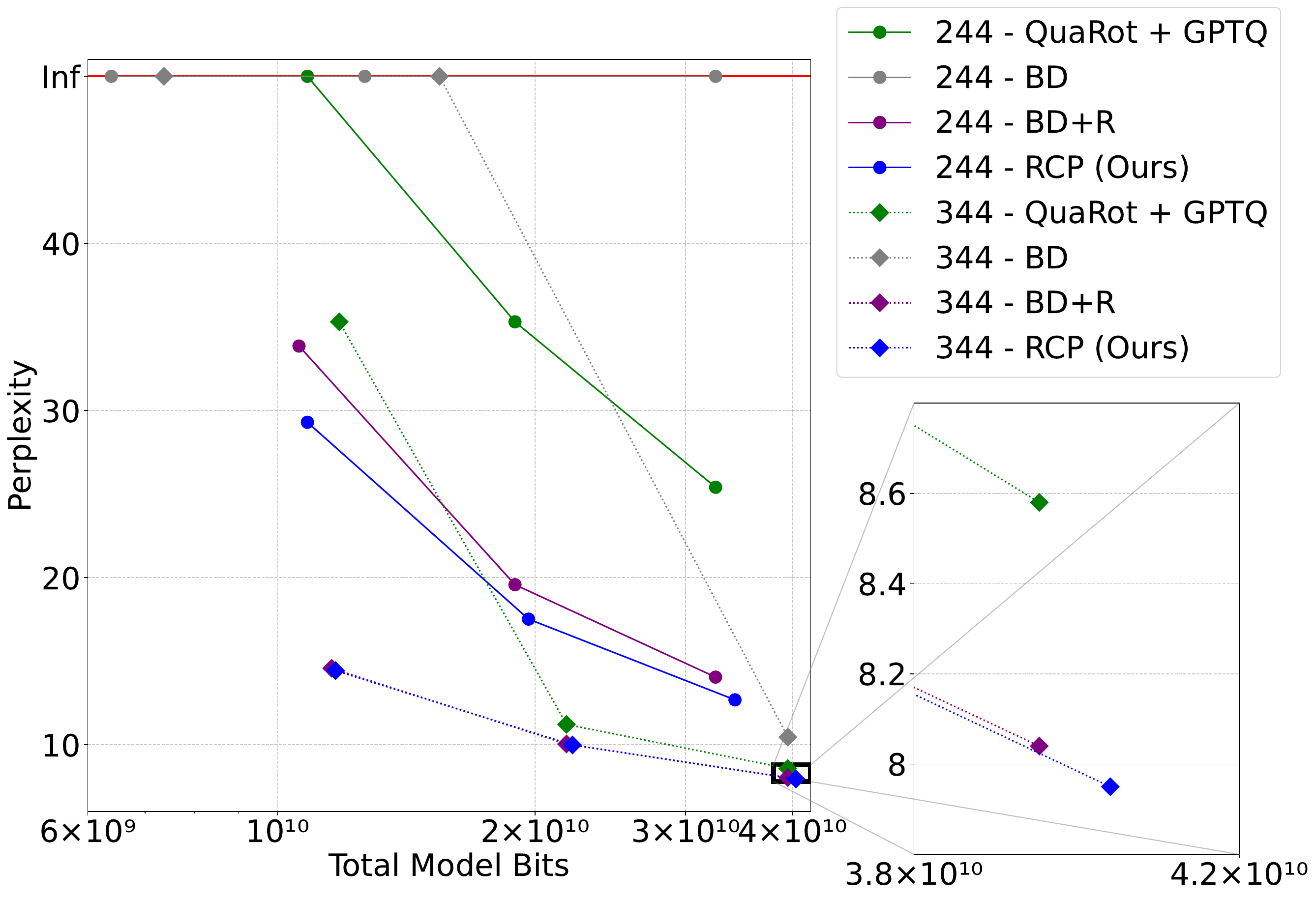} 
    \caption{Bit-Level scaling laws for perplexity for LLaMA-3~\cite{llama3modelcard} (1B, 3B, 8B). }
    \label{fig:pploverview}
\end{figure}
RCP is the first to enable challenging W2A4KV4 and W3A4KV4 quantization on common LLM models, significantly outperforming QuaRot~\cite{quarot} and BitDistiller~\cite{bitdistiller}. Especially, as we present in Fig. \ref{fig:pploverview}, RCP also works on small and mobile-targeted LLaMA-3.2 models~\cite{llama3modelcard} that are harder to quantize. Since there is no available hardware that supports both LUT inference for non-uniform quantization and specialized acceleration for 4-bit activation, we design an accelerated GEMV kernel in CUDA as a proof of concept. Our kernel can reduce the memory footprint up to 5.29 times with a latency lower than the FP16 PyTorch~\cite{pytorch} and INT4 QuaRot implementation.

Our contributions are summarized as follows:
\begin{itemize}
\item{We provide empirical and theoretical analysis on how rotation interacts with weight distribution and poses difficulty on extreme W2A4KV4 quantization.}
\item{To address this issue, we introduce RCP, a quantization algorithm that takes the best from rotation and QAT via LDP, a novel fully learnable non-uniform quantizer.}
\item{We provide extensive experiments to show RCP achieves state-of-the-art W2A4KV4 and W3A4KV4 quantization for the first time.}
\end{itemize}

\section{Preliminaries}
\subsection{Random Rotation for LLM Quantization}
\label{sec:prelim}
\begin{figure}[h!] 
    \centering
    \includegraphics[width=0.8\columnwidth]{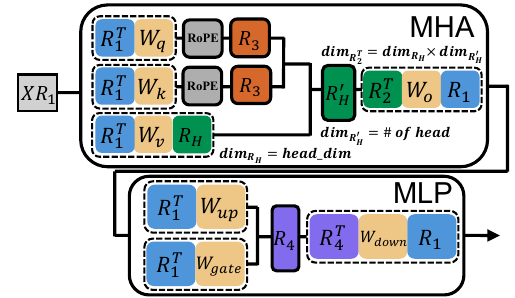} 
    \caption{A diagram of the rotation process in a transformer decoder layer. }
    \label{fig:rotation}
\end{figure}
QuaRot~\cite{quarot} proposed to apply random rotations while keeping the computational invariance suggested in SliceGPT~\cite{slicegpt}. Random rotation suppresses activation outliers and helps quantization, successfully achieving W4A4KV4 with minimal performance loss.

As in Fig. \ref{fig:rotation}, $R_1$ rotates each decoder layer’s inputs and outputs, with its inverse ($R_1^T$) fused into adjacent weights. $R_2$ and $R_4$ are required for online rotation of the MHA and FFN intermediates, respectively. We factorize $R_2$ into two orthogonal matrices—$R_H$ for the V projection and $R_H'$ for the self‐attention activation—and then apply $R_2^T$ to the out‐projection. Finally, $R_3$ rotates Q and K vectors after RoPE, enabling compression of the KV cache without altering attention outputs.



\subsection{Asymmetric Integer Quantizer}

The commonly used asymmetric integer quantization function is defined in Eqn. \ref{eqn:asymmetric}.
\begin{equation}\label{eqn:asymmetric}
\small
\begin{split}
    Q(\mathrm{\mathbf{W}}) = \mathrm{clip}(\lfloor \frac{\mathbf{W}}{s} \rceil + z, 0, 2^{N}-1),\\
    \mathrm{where}\,\, h=\max(\mathbf{W})-\min(\mathbf{W} ),\; s=\frac{h}{2^N-1}\\ 
    z=-\lfloor \frac{\min(\mathbf{W})}{h} \rceil,
\end{split}
\end{equation}
where $N$ is the number of bits, $h$ is the quantization range, $s$ is the step size, and $z$ is the zero-point. This general formulation is applicable to various settings, including per-tensor, per-channel, and group-wise quantization, via adapting the computation of $h$, $s$, and $z$.


\section{Motivation}

%

In this section, we confirm the difficulty of W2A4KV4 quantization by empirical evaluations and justify our key design of non-uniform weight quantizer (in Section \ref{sec:ldp_quant}) through theoretical analysis on the effect of the rotation technique on weight distribution.
\begin{table}[h!]
\centering
\setlength{\tabcolsep}{4pt}
\renewcommand{\arraystretch}{0.8}
\resizebox{0.8\columnwidth}{!}{%
\begin{tabular}{c|c|ccc}
\toprule[1pt]
\raisebox{-2.5ex}{\textbf{Method}} & \raisebox{-2.5ex}{\textbf{$R$}} & \textbf{Language} & \multicolumn{2}{c}{\textbf{Reasoning}} \\
\cmidrule(lr){3-3} \cmidrule(lr){4-5}
& & \textbf{WikiText2$^\downarrow$} & \textbf{Coding$^\uparrow$} & \textbf{Math$^\uparrow$} \\
\midrule
QuaRot & \checkmark & 12772.03 & 0 & 0.002 \\ 
BitDistiller &  & 17.40 & 3.5 & 5.39\\ 
BitDistiller & \checkmark & 8.93 & 6.09&  0.16\\ 
RCP & \checkmark & \textbf{8.31} & \textbf{23.20} & \textbf{40.16} \\
\bottomrule[1pt]
\end{tabular}
}
\caption{Evaluation results on WikiText2, HumanEval, and GSM8K under W2A4KV4. The evaluations are conducted using LLaMA-2 7B for WikiText2 (perplexity), WizardCoder 7B for HumanEval (pass@1), and MetaMath 7B for GSM8K (pass@1). The column $R$ indicates whether rotation is applied.}
\vspace{-0.3cm}
\label{tab:comparison_baseline244}
\end{table}
\vspace{-0.3cm}
\paragraph{Existing algorithms can fail on W2A4KV4}
As shown in Table~\ref{tab:comparison_baseline244}, we first observe that QuaRot and BitDistiller fail under W2A4KV4, particularly on language modeling and reasoning tasks. This demonstrates their limitations: QuaRot effectively mitigates activation outliers but fails to handle extreme low-bit weight quantization. BitDistiller is able to address weight quantization but remains vulnerable to the activation outliers.

Naturally, we conceptualized combining rotation and QAT approaches and conducted experiments with all rotation matrices applied to the BitDistiller implementation. As indicated in the "BitDistiller w/ Rotation" rows in Table~\ref{tab:comparison_baseline244}, language modeling performance was recovered to some extent; however, reasoning capabilities remained non-deployable.

\begin{figure}[t!] 
    \centering
    \includegraphics[width=\columnwidth]{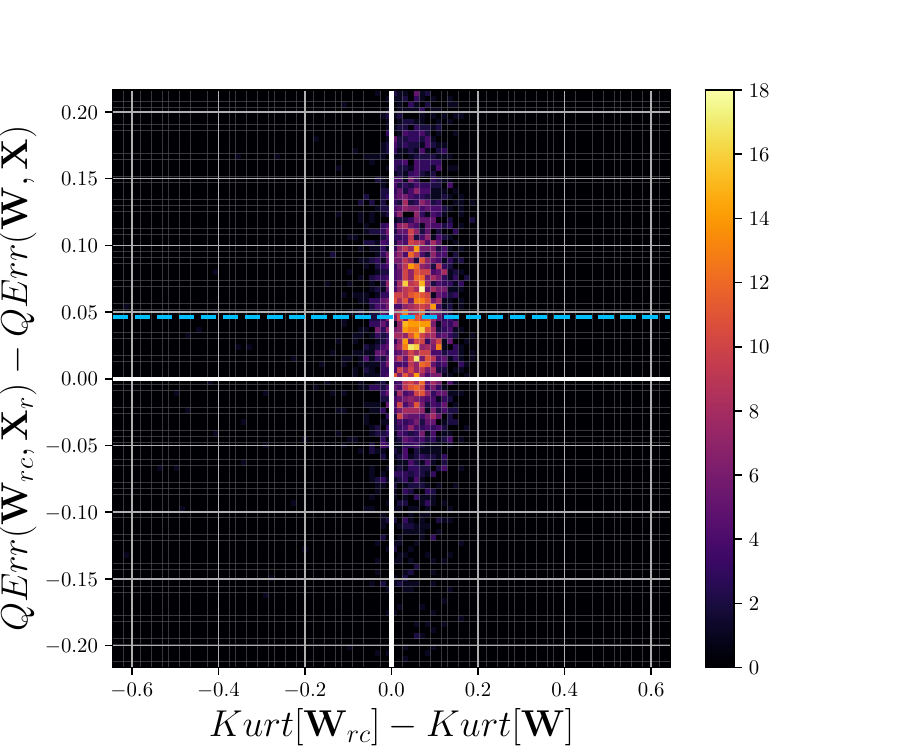} 
    \caption{A two-dimensional histogram comparing the increase in output activation's quantization error and the increase in the averaged group-wise weight kurtosis. The last down-projection layer of the LLaMA-2-7B model is used. The bright blue horizontal line indicates the average increase of the quantization error. We use mean absolute error $QErr(\mathbf{W}, \mathbf{X})=|\mathbf{X}(Q(\mathbf{W})-\mathbf{W})|$ and $\mathbf{W}_{rc}$ follows the definition in Eqn. \ref{eqn:clipinit}.}
    \label{fig:qerr_vs_kurt}
\end{figure}

\paragraph{Relation between rotation and non-uniformity}\label{sec:kurt_analysis}

To explain why such a naive application of rotation to QAT fails, we first theoretically analyze how rotation affects the excess kurtosis of the weight distribution. The excess kurtosis is the shifted fourth standardized moment defined as follows:
\begin{equation}
{Kurt}(X)=\frac{\mu_4}{\sigma^4} - 3,
\end{equation}
where $\mu_4$ is the fourth moment and $\sigma$ is the standard deviation. Larger excess kurtosis indicates a distribution 1) contains numerous outliers and/or 2) is more peaked around its center (i.e., more \textbf{non-uniform}, which is hard to quantize).

Our claim is that the Hadamard matrix (used as rotation) increases the excess kurtosis of a platykurtic distribution\footnote{A distribution $X$ is platykurtic when $Kurt(X) < 0$}, which we empirically observed to be true for most of the LLM weights.
\begin{lemma} \label{lemma:hadamard_kurtosis}
Let $\mathbf{X} = (X_1, X_2, \ldots, X_n)^T$ be a random vector whose components are i.i.d. with finite fourth moment $\mu_4$, mean $\mu$, variance $\sigma^2$, and negative excess kurtosis ($\text{Kurt}(X_i) < 0$). Let $\mathbf{H}_n$ denote the normalized $n \times n$ Hadamard matrix with elements $\pm \frac{1}{\sqrt{n}}$. Then, for the transformed vector $\mathbf{Y} = \mathbf{H}_n\mathbf{X}$, the following holds:
\begin{equation*}
\text{Kurt}(Y_i) > \text{Kurt}(X_i)
\end{equation*}
for all $i \in \{1, 2, \ldots, n\}$.
\end{lemma}

\begin{proof}
The proof can be found in Appendix \ref{sec:lemma1_proof}.
\end{proof}
Since it is well known that the Hadamard matrix is highly effective at eliminating outliers~\cite{quarot,quip,quip-sharp}, rotation is increasing the non-uniformity of the weight distributions. As presented in Fig. \ref{fig:qerr_vs_kurt}, we studied an empirical relation between the increase in the quantization error of output activation and the excess kurtosis of weight, after applying rotation. Clearly, the quantization error is generally enlarged when the excess kurtosis is increased by rotation. See Appendix \ref{sec:qerr_vs_kurt_more_results} for details and further discussion.

\section{Methodology}

\begin{figure*}[htbp]
    \centering
    \begin{subfigure}{0.6\textwidth}
        \includegraphics[width=\linewidth]{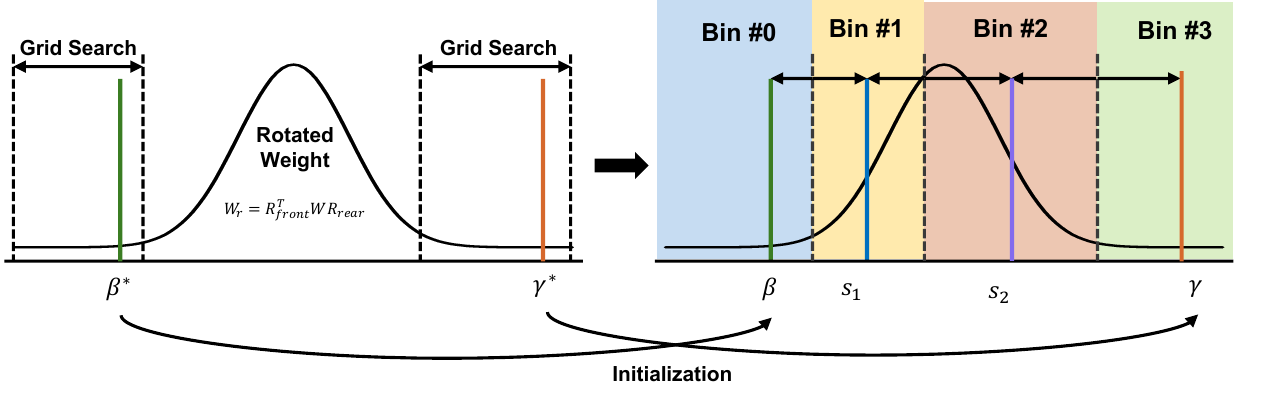}
        \caption{RCP first applies a Hadamard-based rotation to the weight and determines initial clipping values via grid search. The weight is then partitioned into 4 bins with learnable parameters.}
        \label{fig:clip_quantize}
    \end{subfigure}
    \hfill
    \begin{subfigure}{0.35\textwidth}
        \includegraphics[width=\linewidth]{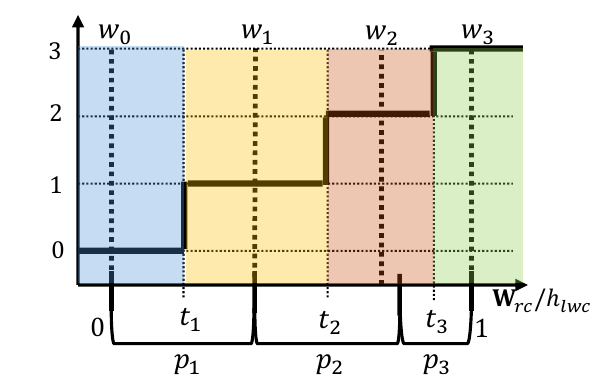}
        \caption{A diagram of Learnable Direct Partitioning.}
        \label{fig:LDP}
    \end{subfigure}
    \caption{Illustration of Learnable Direct Partitioning (LDP) with rotation-aware clipping. }
    \label{fig:main}
\end{figure*}
Our proposition is a QAT algorithm named \textbf{R}otate, \textbf{C}lip, and \textbf{P}artition (\textbf{RCP}) that combines the idea of random rotation with our novel \textbf{L}earnable \textbf{D}irect \textbf{P}artitioning (\textbf{LDP}) quantizer.

Overall, RCP is a self-knowledge distillation~\cite{kd} algorithm that solves the following optimization problem:
\begin{equation}\label{eqn:main_objective}
\small
\begin{split}
\minimize_{\Theta_S} \mathbb{E}_{(x,y)\sim\mathbb{D}}\left[ \mathcal{D}_{CAKLD}(P_{\Theta_T}||P_{\Theta_S}) \right], \\
\mathcal{D}_{CAKLD}(P_{\Theta_T}||P_{\Theta_S}) = \alpha D_{KL}(P_{\Theta_S}||P_{\Theta_T})\\
+ (1-\alpha)D_{KL}(P_{\Theta_T}||P_{\Theta_S}),
\end{split}
\end{equation}
where $\Theta_T$ is the frozen full-precision teacher model, $\Theta_S$ is the student model quantized with LDP, $P_\Theta$ is the logit distribution produced by a model $\Theta$, $\mathbb{D}$ is the training dataset containing pairs of input text $x$ and label text $y$. The $\mathcal{D}_{CAKLD}$ is the confidence-aware KL divergence loss adopted from BitDistiller~\cite{bitdistiller} with the empirical confidence $\alpha$ measured on $\Theta_T$.


\label{sec:method}

\subsection{Rotate: Applying Hadamard Transforms}
The first step of our method is to apply (randomized) Hadamard transforms to model weights, following rotation-based PTQ algorithms~\cite{quarot,spinquant,duquant}. We formulate this procedure as follows:
\begin{equation}
\mathrm{\mathbf{W}}_r=\mathbf{R}^T_{front}\mathrm{\mathbf{W}}\mathbf{R}_{rear},
\end{equation}
where $\mathbf{R}^T_{front}$ and $\mathbf{R}_{rear}$ are Hadamard matrices multiplied to the front and rear side of a model weight $\mathrm{\mathbf{W}}$, respectively. The choice of $\mathbf{R}^T_{front}$ and $\mathbf{R}_{rear}$ can be identified in Fig. \ref{fig:rotation}.

Note that the $\mathrm{\mathbf{W}}_r$ is pre-computed before any optimization to increase memory efficiency and better QAT performance. For further details and ablation, see Appendix \ref{sec:factorized_rotation}.

\subsection{Clip: Learnable Clipping with Grid-Search Initialization}
Clipping is an essential technique to limit quantization range via clamping out extreme values from the maximum and minimum sides of the model  weights~\cite{awq,omniquant,pact,lsq}. 

OmniQuant~\cite{omniquant} introduces learnable weight clipping (LWC) to dynamically determine the optimal quantization range by modifying the static quantization range $h$ in Eqn. \ref{eqn:asymmetric} as follows:
\begin{equation}
\small
h_{lwc}=\sigma(\gamma)\mathrm{max}(\mathrm{\mathbf{W}_{r}})-\sigma(\beta)\mathrm{min}(\mathrm{\mathbf{W}_{r}}),
\end{equation}
where $\beta$ and $\gamma$ are learnable parameters allocated for each quantization unit and $\sigma$ is the sigmoid function.

To enhance stability of RCP, we find the initial point of the clipping parameters $\beta^*,\gamma^*$ in a rotation-aware manner, based on the grid-search strategy~\cite{awq} that solves the following problem on a small calibration dataset $\mathcal{D}_{cal}$:
\begin{equation} \label{eqn:clipinit}
\small
\begin{split}
\beta^*,\gamma^*=\operatorname{argmin}_{\beta, \gamma} \|Q(\mathrm{\mathbf{W}}_{rc})\textbf{X}_{r} - \mathrm{\mathbf{W}}_{rc}\textbf{X}_{r}\|^2,\\
\mathrm{\mathbf{W}}_{rc}=\mathrm{clip}(\mathrm{\mathbf{W}}_{r},\sigma(\beta)\mathrm{min}(\mathrm{\mathbf{W}_r}),\sigma(\gamma)\mathrm{max}(\mathrm{\mathbf{W}_r})),
\end{split}
\end{equation}
where $Q$ is the quantization function defined in Eqn. \ref{eqn:asymmetric}, $\mathrm{\mathbf{W}}_{rc}$ is the \textit{rotated} and \textit{clipped} weight, and $\mathrm{\mathbf{X}}_R$ is the \textit{rotated} activation. In subsequent procedures, $\beta$ and $\gamma$ are learned via backpropagation, constantly searching for optimal dynamic quantization range on given data and updated model weights.

\subsection{Partition: Learnable Direct Partitioning}
The main goal of this work is to design a method to realize a non-uniform integer quantizer that \textit{learns from data}. Prior arts such as N2UQ~\cite{nu2u} and LLT~\cite{llt} combine non-uniform quantization and uniform dequantization schemes to benefit from increased representational capability while being hardware friendly, however, we find this scheme results in suboptimal performance. Instead, LDP performs both steps in a non-uniform fashion to minimize the impact of errors from extreme weight quantization.

\paragraph{Non-uniform quantization via partitioning} \label{sec:ldp_quant}
The core idea of LDP is to \textit{partition} the dynamic quantization range in a differentiable way by introducing two learnable parameters $s_1$ and $s_2$ per quantization unit. By applying sigmoid function to them, LDP directly splits $h_{lwc}$ into three partitions:
\begin{equation}\label{eqn:partition}
\small
\begin{split}
p_1=\sigma(s_1), \\
p_2=(1-p_1)\sigma(s_2), \\
p_3=(1-p_1)(1-\sigma(s_2)),
\end{split}
\end{equation}
where $s_1$ takes the left $\sigma(s_1)*100$\% of $h_{lwc}$, $s_2$ takes the left $\sigma(s_2)*100$\% of the remaining range $(1-p_1)h_{lwc}$, and the last partition is determined trivially.

We set the quantization grid $\{t_i|i\in\{1,2,\dots,2^N-1\}\}$ at the center of each partition as follows:
\begin{equation}
\small
t_i=t_{i-1}+\frac{p_{i-1}+p_i}{2},\; \mathrm{where}\; t_1=p_1/2.
\end{equation}
This obtains the quantization function of LDP as follows:
\begin{equation}\label{eqn:quant}
\small
Q_{LDP}(\textbf{W}_{rc}) = \text{u}(\frac{\textbf{W}_{rc}}{h_{lwc}}-t_1) + \text{u}(\frac{\textbf{W}_{rc}}{h_{lwc}}-t_2) + \text{u}(\frac{\textbf{W}_{rc}}{h_{lwc}}-t_3),
\end{equation}
where $u(x)$ is the step function.

We rationalize this formulation in three points: 1) the dynamic range $h_{lwc}$ is seamlessly filled out since $\sum_{i=1}^3p_i=1$ is guaranteed; 2) each partition is constrained between 0\% and 100\% as the sigmoid re-parametrization ensures each $p_i\in[0,1]$; 3) no matter how the partitioning parameters are updated, the ordering of the partitions stays the same by the construction of $p_i$.

The initialization of $s_1$ and $s_2$ is set to $\sigma^{-1}(1/3)$ and $\sigma^{-1}(1/2)$, respectively (i.e., the dynamic range is uniformly partitioned at the beginning). Technically, the grid-search strategy~\cite{awq} can also be employed to jointly find the optimal partitioning parameters; however, the computational cost will grow exponentially since we have to iterate over a four-dimensional optimization loop (two for LWC and two for LDP).

\paragraph{Non-uniform dequantization} \label{sec:ldp_dequant}
The overall design of the quantization function in Eqn. \ref{eqn:quant} is imported, and the dequantization function of LDP is given by:
\begin{equation}\label{eqn:dequant}
\small
\begin{split}
DQ_{LDP}(\mathrm{\mathbf{W}}_{rc}) =& \sigma(\beta)\text{min}(\textbf{W}_{rc}) \\
+ h_{lwc}\biggl( &\text{u}(\frac{\textbf{W}_{rc}}{h_{lwc}}-t_1)(w_1-w_0) \\
+ &\text{u}(\frac{\textbf{W}_{rc}}{h_{lwc}}-t_2)(w_2-w_1) \\
+ &\text{u}(\frac{\textbf{W}_{rc}}{h_{lwc}}-t_3)(w_3-w_2) \biggr),
\end{split}
\end{equation}
where the dequantization grid $\{w_i|i\in\{0,1,2,3\}\}$ is additionally introduced in Eqn. \ref{eqn:dequantgrid}.
\begin{equation}\label{eqn:dequantgrid}
\small
w_i=\begin{cases}
    0, & \text{if } i=0 \\
    \frac{t_i+t_{i+1}}{2}, & \text{if } 0 < i < 3 \\
    1, & \text{if } i=3.
\end{cases}
\end{equation}
This means that the full-precision weight elements whose normalized value is smaller than the first quantization grid (i.e., $\mathrm{W}/h_{lwc}<t_1$) are mapped to the minimum possible value in the dynamic range $\sigma(\beta)\mathrm{min}(\mathrm{\mathbf{W}_{rc}})$. Likewise, the elements that satisfy $\mathrm{W}/h_{lwc}>t_3$ are mapped to the maximum value $\sigma(\gamma)\mathrm{max}(\mathrm{\mathbf{W}_{rc}})$\footnote{Since $\sigma(\beta)\mathrm{min}(\mathrm{\mathbf{W}_{rc}})+h_{lwc}=\sigma(\gamma)\mathrm{max}(\mathrm{\mathbf{W}_{rc}})$}. The others in the middle are mapped to the center of the second and third quantization bin, realizing non-uniform dequantization.

Finally, when computing the loss function in Eqn. \ref{eqn:main_objective}, each weight $\mathbf{\theta}_S\in\Theta_S$ is fake-quantized by Eqn. \ref{eqn:dequant} as $\mathbf{\theta}_S\leftarrow DQ_{LDP}(\mathbf{\theta}_s)$.
We note that during the fake quantization, every step function $u(\cdot)$ is applied with the straight-through estimator so that every parameter (including LLM weights, clipping, and partitioning parameters) can be updated via backpropagation.

\paragraph{Application of LDP on NF3 format}
We apply not only 2-bit integer weight quantization but also 3-bit quantization using the asymmetric NF format of AFPQ~\cite{afpq} where separate scale values are computed for the negative and positive weights ($s_{neg}=\text{max}(|\textbf{W}_{rc,neg}|)$, $s_{pos}=\text{max}(\textbf{W}_{rc,pos})$). Although shown to be effective, such an NF3 quantizer can lead to suboptimal performance when the distribution is not zero-centered. Therefore, we make a further improvement by applying the LDP to this situation.

The idea is to employ the same learnable clipping parameters ($\beta$, $\gamma$) to obtain the dynamic quantization range $h_{lwc}$ and one partitioning parameter $s_1$ to express the learnable center point as $c = \sigma(\beta)\text{min}(\textbf{W}) + h \cdot \sigma(s_1)$. Then, the two scale values are updated as follows:
\begin{equation}\label{eqn:nf3scales}
\small
\begin{split}
s_{neg} = |c-\sigma(\beta)\text{min}(\textbf{W}_{rc})|, \\
s_{pos} = |\sigma(\gamma)\text{max}(\textbf{W}_{rc}) - c|,
\end{split}
\end{equation}
\normalsize
and the quantization process is derived as follows:
\begin{equation}\label{eqn:nf3}
\textbf{W}_q = \begin{cases}
\lfloor \frac{\textbf{W}_{rc}-c}{s_{pos}} \rceil, & \text{if } \textbf{W}_{rc} > c \\
\lfloor \frac{\textbf{W}_{rc}-c}{s_{neg}} \rceil, & \text{otherwise.}
\end{cases}
\end{equation}
The dequantization is done simply by multiplying the scales to $\textbf{W}_q$ and adding $c$.

\begin{figure*}[t!] 
    \centering
    \includegraphics[width=0.85\textwidth]{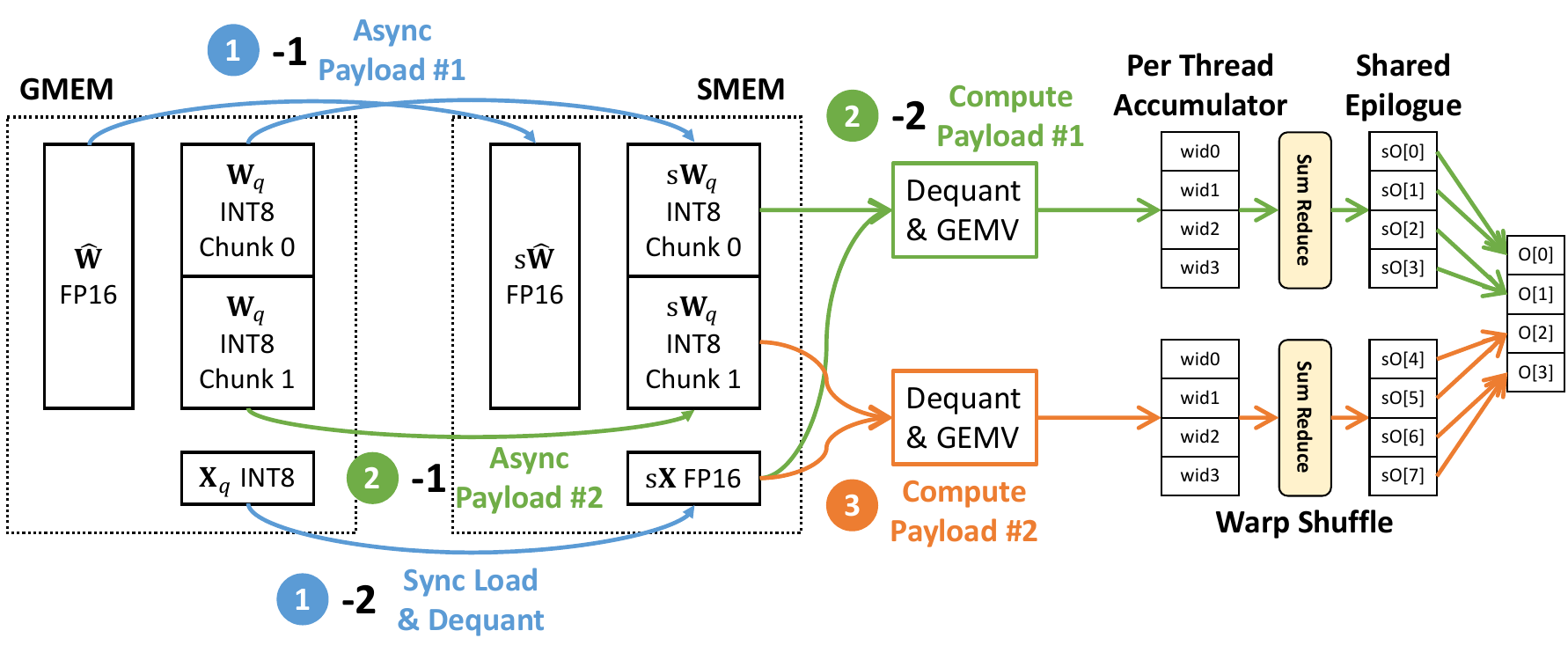} 
    \caption{An overview of our GPU GEMV kernel with data path along memory hierarchy, pipelining, and epilogue concisely illustrated. $\texttt{wid}$ is the warp index and the per-thread accumulator is simplified (warp lane dimension is not shown).}
    \vspace{-0.5cm}
    \label{fig:gemv}
\end{figure*}

\subsection{Look-up Table (LUT) Inference for Non-uniform W2A4 GEMV}
\label{sec:W2A4Kernel}

To implement the non-uniform W2A4 inference on modern GPUs, LUT-based GEMM and GEMV kernels are designed. The quantized INT2 weights $\mathbf{W_q}$ and the FP16 dequantization LUT $\hat{\mathbf{W}}$ are pre-computed from LDP using Eqn. \ref{eqn:quant} and \ref{eqn:dequant} without incurring any runtime overhead as follows:
\begin{equation}\label{eqn:deqgrid}
\begin{split}
\mathbf{W}_q = Q_{LDP}(\mathbf{W}_{rc}), \\
\hat{\textbf{W}} = \{\hat{\text{W}}_0, \hat{\text{W}}_1, \hat{\text{W}}_2, \hat{\text{W}}_3\} \\
\mathrm{where}\; \hat{\text{W}}_i=\sigma(\beta)\mathrm{min}(\mathbf{W}_{rc}) + h_{lwc}\cdot w_i.
\end{split}
\end{equation}

However, designing such kernels poses a big challenge. First, efficient INT tensor cores cannot be utilized since accumulating the multiplication results in INT quantized space makes it impossible to dequantize the weights back to correct non-uniform real values in the LUT $\hat{\textbf{W}}$. Second, both weights and activations must undergo online dequantization to support dynamic quantization, which adds a large amount of computation overhead.

Therefore, we focus on designing GEMV kernel for LUT decoding predominantly bounded by memory bandwidth, which is ideal for featuring the advantage of extreme W2A4KV4 compression. We report our exploratory results on GEMM design in Section \ref{sec:gemm}.

\paragraph{Fast GEMV via Latency Hiding}\label{sec:gemv_datapath}
We define the input channel dimension as C, the output channel dimension as H, and the number of groups per channel as N. The quantized activation tensor $\textbf{X}_q$ has a shape of 1 × C/2 and is INT8, with each element holding two INT4 activation elements. The activation scale $\text{S}$ is an FP16 scalar. The quantized weight tensor $\textbf{W}_q$ has a shape of H × C/4 and is INT8, with each element holding four UINT2 weights. The dequantization grid $\hat{\textbf{W}}$ has a shape of H × N$\cdot$4 and is FP16. The output activation $\textbf{O}$ is an FP16 tensor of shape 1 × H.

As demonstrated in Fig. \ref{fig:gemv}, we store the dequantized input activation $\text{s}\textbf{X}$ (1 × C, FP16), the quantized weight tile $\text{s}\textbf{W}_q$ (BH × C/4, INT8), the corresponding dequantization grid tile $\text{s}\hat{\textbf{W}}$ (BH × N$\cdot$4, FP16), and a shared output array $\text{s}\textbf{O}$ (1 × 8, FP16) in shared memory.

To make our kernel efficient via latency hiding, we design a pipelining strategy where a thread block handles a half of the output elements (BH/2) and iterates twice. At the beginning, an asynchronous copy of $\hat{\textbf{W}}$ and the first $\textbf{W}_q$ chunk (of size BH/2 × C/4) is issued using \texttt{cp.async} instruction (\textbf{1-1} in Fig. \ref{fig:gemv}). Simultaneously, $\textbf{X}_q$ is synchronously loaded from global memory and dequantized to be stored into $\text{s}\textbf{X}$ (\textbf{1-2}), overlapping activation dequantization latency with loading the first weight chunk.

Subsequently, while we bring in the second $\textbf{W}_q$ chunk using \texttt{cp.async} (\textbf{2-1}), we perform dequantization, inner product, and warp reduce on the first $\textbf{W}_q$ chunk at the same time (\textbf{2-2}), thereby hiding the second chunk loading latency with computation of the first chunk. Finally, the computation on the second chunk is performed (\textbf{3}) and the shared output array is reduced once more if necessary.

Additional details (e.g., dequantization implementation, shared output) not mentioned here are provided in Section \ref{sec:gemvdetail}. 


\begin{table*}[htbp]
    \centering
    \setlength{\tabcolsep}{5pt}
    \renewcommand{\arraystretch}{1.5}
    \resizebox{\textwidth}{!}{%
    {\Huge
    \begin{tabular}{cccc|ccc|ccc|ccc|ccc|ccc|ccc}
        \toprule[4pt]
       \raisebox{-4ex}{ \textbf{\#Bits \(\text{(W-A-KV)}\)}} & \multicolumn{3}{c}{ \raisebox{-2ex}{\textbf{Configuration}} }& \multicolumn{3}{c}{ \raisebox{-2ex}{\textbf{LLAMA-1 7B}}} & \multicolumn{3}{c} {\raisebox{-2ex}{\textbf{LLAMA-2 7B}}} & \multicolumn{3}{c} {\raisebox{-2ex}{\textbf{LLAMA-2 13B}}} &\multicolumn{3}{c}{ \raisebox{-2ex}{\textbf{LLAMA-3 8B}}} & \multicolumn{3}{c}{ \raisebox{-2ex}{\textbf{LLAMA-3.2 1B}}} & \multicolumn{3}{c}{ \raisebox{-2ex}{\textbf{LLAMA-3.2 3B}}}   \\
        \cmidrule(lr){2-4} \cmidrule(lr){5-7} \cmidrule(lr){8-10} \cmidrule(lr){11-13} \cmidrule(lr){14-16} \cmidrule(lr){17-19} \cmidrule(lr){20-22}
        & \textbf{Method} & \textbf{Rotation} & \textbf{LDP} & \textbf{MMLU} & \textbf{0-shot$^\dagger$} & \textbf{Wiki$^\downarrow$} & \textbf{MMLU} & \textbf{0-shot$^\dagger$} & \textbf{Wiki$^\downarrow$} & \textbf{MMLU} & \textbf{0-shot$^\dagger$} & \textbf{Wiki$^\downarrow$} & \textbf{MMLU} & \textbf{0-shot$^\dagger$} & \textbf{Wiki$^\downarrow$} & \textbf{MMLU} & \textbf{0-shot$^\dagger$} & \textbf{Wiki$^\downarrow$} &  \textbf{MMLU} & \textbf{0-shot$^\dagger$} & \textbf{Wiki$^\downarrow$}   \\
        \midrule
        \multirow{1}{*}{16-16-16} & BF16 &  &  & 35.10 & 68.40 & 5.68 &46.45 & 61.67 & 5.47 & 55.54& 63.02 &4.88 & 68.40 & 72.93 & 6.10 & 32.20 & 58.90 & 13.40 &58.00 &65.30 & 10.70 \\
        \midrule
        \multirow{3}{*}{2-4-16} 
        & BitDistiller &  &  & 25.88 & 42.56 & 23.19 & 26.24 & 43.36 & 16.47 &26.05 &39.66 & 23.16& 23.11 & 39.46&Inf &25.00&36.82 & Inf & 24.41 &37.89 &  Inf \\
        & BitDistiller & \checkmark &  & 26.75 & 52.28 & 8.79 & 26.04 & \textbf{51.49} & 8.93 &29.97 & 48.48& 7.55& 29.80 & 50.59 & 13.68 & 25.00 & 41.08 & 31.32 & 29.60 & 45.29 & 18.79    \\
        & RCP & \checkmark & \checkmark & \textbf{27.34} & \textbf{52.29} & \textbf{8.28} & \textbf{28.04} & 51.10 & \textbf{8.18} & \textbf{37.27} &\textbf{51.71} & \textbf{7.27} & \textbf{31.87} & \textbf{50.86} & \textbf{12.48} & \textbf{26.30} & \textbf{41.35} & \textbf{27.46} & \textbf{31.40} & \textbf{45.71} & \textbf{16.96 }  \\
        \midrule
        \multirow{3}{*}{2-4-4} 
        & BitDistiller &   &   & 24.45 & 43.08 & 19.98 & 26.59 & 44.93 & 17.40 &24.72 &36.73 &32.43 & 23.29 & 39.75 & Inf & 24.66 & 37.55 & Inf & 24.26 & 37.26 & Inf   \\
        & BitDistiller & \checkmark &   & 26.98 & 52.21 & 8.92 & 26.41 & 51.10 & 8.93 & 24.18 &43.55 & 11.45& 29.66 & 49.80 & 14.05 & 24.74 & 40.77 & 33.86 & \textbf{31.44} & 44.26 & 19.58   \\
        & RCP & \checkmark & \checkmark & \textbf{27.34} & \textbf{52.29} & \textbf{8.28} & \textbf{26.92} & \textbf{51.22} & \textbf{8.31} &\textbf{35.49} &\textbf{48.18} &\textbf{7.95} & \textbf{31.01} & \textbf{50.41} & \textbf{12.69} & \textbf{25.62} & \textbf{41.80} & \textbf{29.30} & 30.33 & \textbf{45.56} & \textbf{17.52}   \\
        \midrule
        \multirow{3}{*}{3-4-16} 
        & BitDistiller &   &   & 26.88 & 55.68 & 7.47 & 31.72 & 56.15 & 7.04 &42.68 &54.59 &6.99 & 42.24 & 55.39 & 10.19 & 26.06 & 37.53 & Inf & 25.22 & 37.32 & Inf   \\
        & BitDistiller & \checkmark &  & 28.70 & 58.52 & 6.44 & 34.30 & 59.28 & 6.25 &46.91 &58.99 &5.62 & 54.16 & 61.06 & 7.92 & 26.45 & 47.88 & 13.75 & \textbf{47.34} & 55.66 & 9.82   \\
        & RCP & \checkmark & \checkmark & \textbf{29.46} & \textbf{59.39} & \textbf{6.39} & \textbf{37.33} & \textbf{59.74} & \textbf{6.23} & \textbf{50.84}& \textbf{60.52}& \textbf{5.49}& \textbf{55.33} & \textbf{61.53} & \textbf{7.80} & \textbf{27.77} & \textbf{48.18} & \textbf{13.68} & 47.31 & \textbf{55.87} & \textbf{9.74}   \\
        \midrule
        \multirow{3}{*}{3-4-4} 
        & BitDistiller &   &   & 27.04 & 56.05 & 7.54 & 30.19 & 55.51 & 7.15 & 40.58&54.57 & 9.02& 40.70 & 56.35 & 10.46 & 25.48 & 38.75 & Inf & 25.91 & 37.27 & Inf    \\
        & BitDistiller & \checkmark &   & 28.80 & 58.48 & 6.45 & 33.46 & 58.53 & 6.36 &47.86 &58.78 &6.06 & 51.74 & 59.69 & 8.04 & 26.11 & 47.14 & 14.58 & 46.08 & 55.08 & 10.05   \\
        & RCP & \checkmark & \checkmark & \textbf{30.00} & \textbf{58.55} & \textbf{6.39} & \textbf{36.07} & \textbf{59.27} & \textbf{6.33} &\textbf{48.47} &\textbf{58.83} &\textbf{5.57} & \textbf{52.55} & \textbf{61.11} & \textbf{7.95} & \textbf{26.54} & \textbf{47.71} & \textbf{14.44} & \textbf{46.40} & \textbf{55.12} & \textbf{9.99}   \\
        \bottomrule[4pt]
    \end{tabular}
    }
    }
    \caption{Comparison of the perplexity score on WikiText2, MMLU (5s) and averaged accuracy on four Zero-shot Common
Sense Reasoning tasks. We show the perplexity results
>100 by Inf. Full results of Zero-shot tasks are in the Appendix. }
\vspace{-0.1cm}
\label{tab:main_result}
\end{table*}










\section{Experiments}
\label{sec:experiments}

\subsection{Experimental Settings}
\paragraph{Models and Tasks}
We evaluate RCP on LLaMA-1~\cite{llama} 7B, LLaMA-2~\cite{llama2}(7B, 13B), LLaMA-3~\cite{llama3modelcard}(1B, 3B, 8B). Our evaluation of RCP was carried out on PIQA~\cite{piqa}, HellaSwag~\cite{hellaswag}, WinoGrande~\cite{winogrande}, ARC-c~\cite{arc}, MMLU~\cite{mmlu} and LongBench~\cite{longbench}. We use LLM-Humaneval-Benchmarks~\cite{humaneval} and GSM8K~\cite{gsm8k} for reasoning task evaluation. We also report the perplexity score on WikiText2~\cite{wiki2} for our evaluation.
\paragraph{Training Data}
For a fair comparison with our baseline, we use the instruction-tuning data from Alpaca~\cite{alpaca} and the training set of WikiText2 for general language tasks.
For understanding and generating code, we use Evol-Instruct-Code\footnote{https://github.com/nickrosh/evol-teacher}. For math reasoning we use MetaMathQA~\cite{metamathQA}.
\paragraph{Training Configurations}
We set the weight learning rate to 8e-7 for W$l$A4 and 1e-6 for W$l$A4KV4, while the learning rate for LWC and LDP was set to 1e-5. We set the training sequence length to 1024 and the evaluation sequence length to 2048. 


\subsection{Results}
We compare our proposed RCP with the state-of-the-art QAT method, BitDistiller~\cite{bitdistiller}. Details on training cost and implementation are provided in Appendix~\ref{sec:additional_result}.
\paragraph{Language Modeling Tasks}
The results are summarized in Table \ref{tab:main_result}. From the perspective of general language tasks, our method demonstrates the ability to quantize activations and KV-cache under the W2 settings to 4-bit, which was previously unattainable using existing QAT methods. The application of rotation effectively addresses the outlier issues, a common bottleneck in quantization, enabling stable performance even in extremely low-bit quantization scenarios. Furthermore, the addition of LDP improves performance on general language tasks across the board, and generally enhances the accuracy of zero/few-shot tasks, which were not adequately addressed by rotation alone. For example, the addition of LDP contributes to a performance gain from 11.45 to 7.95 on LLaMA-2 13B, demonstrating its effectiveness across model scales.  

\paragraph{Reasoning Tasks}
\begin{table}[htbp]
    \centering
    \setlength{\tabcolsep}{2pt}
    \renewcommand{\arraystretch}{0.8}
    \resizebox{\columnwidth}{!}{%
    {
    \begin{tabular}{cccc|c|c}
        \toprule[2pt]
       \raisebox{-1ex}{ \textbf{\#Bits \(\text{(W-A-KV)}\)}} & \multicolumn{3}{c}{ \raisebox{-0ex}{\textbf{Configuration}} }& \multicolumn{1}{c}{ \raisebox{-0ex}{\textbf{WizardCoder 7B}}} & \multicolumn{1}{c} {\raisebox{-0ex}{\textbf{MetaMath 7B}}} \\
        \cmidrule(lr){2-6} 
        & \textbf{Method} & \textbf{Rotation} & \textbf{LDP} & \textbf{HumanEval} & \textbf{GSM8K}   \\
        \midrule
        \multirow{1}{*}{16-16-16} & BF16  &  &  & 54.88 & 66.41  \\
        \midrule
        \multirow{3}{*}{2-4-16} 
        & BitDistiller &  &  & 2.43 & 0.0 \\
        & BitDistiller & \checkmark &  & 14.63 & 1.25    \\
        & RCP & \checkmark & \checkmark & \textbf{27.44} & \textbf{41.64}   \\
        \midrule
        \multirow{3}{*}{2-4-4} 
        & BitDistiller &   &   & 3.50 & 5.39    \\
        & BitDistiller & \checkmark &   & 6.09 & 0.16   \\
        & RCP & \checkmark & \checkmark & \textbf{23.20} & \textbf{40.16}   \\
        \midrule
        \multirow{3}{*}{3-4-16} 
        & BitDistiller &   &   & 0.0 & 0.0  \\
        & BitDistiller & \checkmark &  & 39.02 & 0.0    \\
        & RCP & \checkmark & \checkmark & \textbf{40.85} & \textbf{54.69}   \\
        \midrule
        \multirow{3}{*}{3-4-4} 
        & BitDistiller &   &   & 0.0 & 0.0    \\
        & BitDistiller & \checkmark &   & 41.46 & 0.0    \\
        & RCP & \checkmark & \checkmark & \textbf{43.29} & \textbf{52.73}  \\
        \bottomrule[2pt]
    \end{tabular}
    }
    }
    \caption{Reasoning task results of RCP on domain-specific LLMs.}
    \label{tab:reason}
    \vspace{-0.1cm}
\end{table}


The results of the reasoning tasks are summarized in Table \ref{tab:reason}. We evaluate reasoning capabilities in the domains of coding and mathematics. 

For the coding domain-specific model, WizardCoder~\cite{luo2023wizardcoder}, BitDistiller failed to offer the functional quantized models in both W3 and W2 settings. 
In our method, applying rotation alone was not effective in W2 settings and recovered some output quality in W3 settings.
By incorporating LDP, we achieved up to a threefold improvement in performance, with accuracy increasing from 6.09\% to 23.20\% under the W2A4KV4.
As shown in Fig. \ref{fig:coding}  with the application of LDP, we were able to produce logically correct code outputs and eliminate repetition of meaningless code generation.

For the mathematical reasoning model, MetaMath~\cite{metamathQA}, the baseline BitDistiller failed to offer functional quantized models while ours with LDP could produce working quantized models. These results highlight the critical role of LDP in enabling proper task performance for reasoning models under extreme low-bit quantization. The output comparison for this task is summarized in Fig. \ref{fig:gsm8k_full}.
\paragraph{Long-Context Benchmarks}

\begin{table}[htbp]
\centering
\renewcommand{\arraystretch}{0.3}
\resizebox{0.7\columnwidth}{!}{%
{
\begin{tabular}{c|c|c}
\toprule[2pt]
\textbf{LLaMA-2 7B} & \textbf{Method} & \textbf{Avg.} \\
\midrule
\multirow{3}{*}{Chat-4k} 
    & BF16         & 32.53 \\
    & BitDistiller & 4.37 \\
    & RCP          & \textbf{19.32} \\
\midrule
\multirow{3}{*}{Instruct-32k} 
    & BF16         & 27.13 \\
    & BitDistiller & 5.16 \\
    & RCP          & \textbf{12.29} \\
\bottomrule[2pt]
\end{tabular}
}
}
\caption{Comparison of LongBench results of RCP under W2A4KV4 across different models and methods.}
\label{tab:longbench_avg}
\end{table}
We conduct experiments on a subset of the LongBench dataset to evaluate the effectiveness of our method under various context lengths. Specifically, we test both LLaMA-2 7B-chat-4k and LLaMA-2 7B-Instruct-32k models across eight benchmark tasks. As shown in Table~\ref{tab:longbench_avg}, our proposed RCP with W2A4KV4 consistently outperforms BitDistiller with W2A4KV4 across all tasks. For instance, on the LLaMA-2 7B-chat-4k model, RCP achieved an average score of 19.32, significantly higher than BitDistiller’s 4.37. Similarly, on the LLaMA-2 7B-Instruct-32k model, RCP yields 12.29 compared to BitDistiller’s 5.16, demonstrating robustness to extended context lengths.  These findings further support the effectiveness of RCP-based quantization in preserving reasoning capability under constrained precision and longer context. The detailed results for each benchmark are presented in Table~\ref{tab:4k-long} and Table~\ref{tab:32k-long}.
\paragraph{Inference}
\begin{table}[h!]
\resizebox{\columnwidth}{!}{%
\centering
\renewcommand{\arraystretch}{0.9} 
\begin{tabular}{l|ccc}
\toprule[2pt]
\textbf{Layer Size} & \textbf{(2048, 2048)} & \textbf{(3072, 3072)} & \textbf{(4096, 4096)} \\ \specialrule{0.5pt}{0pt}{0pt}
\textbf{FP16} & 0.042 & 0.047 & 0.051 \\
\textbf{QuaRot} & 0.077 & 0.057 & 0.078 \\
\textbf{QuaRot+FP16Had} & 0.158 & 0.210 & 0.159 \\
\textbf{QuaRot+FP32Had} & 0.194 & 0.238 & 0.191 \\
\textbf{RCP} & 0.028 & 0.03 & 0.040 \\
\textbf{RCP+FP16Had} & 0.114 & 0.167 & 0.110 \\
\textbf{RCP+FP32Had} & 0.136 & 0.204 & 0.148 \\
\bottomrule[2pt]
\end{tabular}
}

\caption{GEMV latency without activation quantization overhead. The layer size is composed as (input channel, output channel). All latency numbers are in milliseconds. Full results are in the Appendix.}
\label{tab:latency}
\end{table}

\begin{table}[h!]
\resizebox{\columnwidth}{!}{%
\centering
\renewcommand{\arraystretch}{1.2} 
\begin{tabular}{c|ccccc}
\toprule[2pt]
\textbf{} & \textbf{3.2-1B} & \textbf{3.2-3B} & \textbf{1.2-7B} & \textbf{3-8B} & \textbf{2-13B} \\ \specialrule{2pt}{0pt}{0pt}
\textbf{FP16} & 2.47GB & 6.43GB & 13.48GB & 16.06GB & 26.03GB \\ \hline
\textbf{BD W3} & 0.92GB (2.68x) & 1.93GB (3.33x) & 3.16GB (4.26x) & 4.94GB (3.25x) & 5.81GB (4.48x) \\
\textbf{RCP W3} & 1.46GB (1.69x) & 2.77GB (2.32x) & 3.26GB (4.14x) & 5.05GB (3.18x) & 6.01GB (4.33x) \\
\textbf{BD W2} & 0.80GB (3.08x) & 1.58GB (4.06x) & 2.35GB (5.73x) & 4.07GB (3.94x) & 4.22GB (6.17x)\\
\textbf{RCP W2} & 1.35GB (1.82x) & 2.46GB (2.62x) & 2.55GB (5.29x) & 4.28GB (3.75x) & 4.62GB (5.63x) \\
\bottomrule[2pt]
\end{tabular}}
\caption{Memory footprint comparison for different weight precisions. Note that 1.2-7B refers to LLaMA-1 and LLaMA-2.}
\label{tab:memory}
\end{table}
Table \ref{tab:latency} and \ref{tab:memory} present the results for GEMV in terms of latency and memory comsumption. The latency of GEMV, excluding the activation quantization overhead, is faster compared to FP16 and QuaRot~\cite{quarot}. This improvement can be attributed to the lower bit precision, which enhances computational efficiency. Table \ref{tab:memory} measures the peak memory footprint for W2A4 and W3A4. Although RCP incurs memory overhead due to additional parameters per quantization group beyond the BitDistiller, the performance gain from RCP compensates for this cost. For W2A4, a significant reduction on 5.29x in memory footprint was achieved compared to FP16. Note that in the LLaMA-3.2 series, it is necessary to separate the embedding table and head modules to satisfy the invariance arising from their tying. Furthermore, as the size of the embedding table has increased compared to previous models, the compression ratio has decreased accordingly.

The end-to-end inference latency is also measured under the W2-only setting, and the results are presented in Table \ref{tab:W2_inference}. -BF16 is the original WizardCoder 7B, and -RCP is our model. We report prefill, decode, and total latency values. The weight-only version of our GEMV kernel is attached to the TinyChat engine~\cite{awq} and several problems from the HumanEval benchmark where our RCP model provided correct answers are randomly chosen. We merged the non-uniform weight dequantization part of the W2A4 GEMV and the C++ PyTorch linear kernel (which works in BF16) into a single CUDA kernel to reduce launch overhead. The BF16 model used the C++ linear kernel only. In the prefill stage, RCP is slower than BF16 due to online dequantization overhead. In the decode stage, however, 2-bit weight representation with our efficient non-uniform dequantization greatly reduces the memory traffic from global to shared memory, resulting in an average decoding latency reduced by 30.19\%. Combined, RCP can run faster on decoding-oriented tasks.

\begin{table}[h]
\centering
\resizebox{1.0\columnwidth}{!}{
\renewcommand{\arraystretch}{1.2}
\begin{tabular}{l|r|r|r|r|r|r|r|r}
\toprule[2pt]
\textbf{Problem } & \textbf{\#1} &\textbf{\#2} & \textbf{\#3} & \textbf{\#4} & \textbf{\#5} &  \textbf{\#6} &  \textbf{\#7} & \textbf{Avg.}\\
\hline
Prefill-BF16     & 1.37 & 0.17 & 0.18 & 0.18 & 0.16 & 0.16 & 0.17 & 0.34 \\
Prefill-RCP    & 1.29 & 0.23 & 0.24 & 0.25 & 0.23 & 0.19 & 0.22 & 0.38 \\
Decode-BF16    & 13.18 & 13.35 & 13.45 & 13.08 & 13.25 & 13.83 & 13.52 & 13.38 \\
Decode-RCP     & 9.26 & 9.32 & 9.47 & 9.15 & 9.08 & 9.81 & 9.27 & 9.34 \\
Total-BF16    & 14.55 & 13.52 & 13.63 & 13.26 & 13.41 & 13.99 & 13.69 & 13.72 \\
Total-RCP    & 10.55 & 9.55 & 9.71 & 9.40 & 9.31 & 10.00 & 9.49 & 9.72 \\
\bottomrule[2pt]
\end{tabular}
}
\caption{Comparison of end-to-end time per inference phase across different problems between BF16 and W2-only (LDP) quantization for WizardCoder 7B. All values are measured in ms/token.}
\label{tab:W2_inference}
\end{table}
\paragraph{Training Cost}
\begin{table}[h]
\centering
\resizebox{\columnwidth}{!}{
\begin{tabular}{llccccc}
\toprule[2pt]
\textbf{Method} & \textbf{Metric} & \textbf{1B} & \textbf{3B} & \textbf{7B} & \textbf{8B} & \textbf{13B} \\
\midrule
\multirow{4}{*}{BitDistiller} & VRAM (GB) & 35.1 & 42.2 & 32.0 & 77.4 & 130.2 \\
                              & Time (h)  & 64.0 & 64.8 & 68.4 & 93.6 & 29.6 \\
                              & Epoch & 16 & 8 & 8 & 8 & 8 \\
                              & Batch & 4 & 4 & 8 & 4 & 32 \\ \hline
\multirow{4}{*}{RCP}          & VRAM (GB) & 35.3 & 42.9 &33.1 & 78.2 & 132.3 \\
                              & Time (h)  & 67.2 & 69.6 & 73.3 & 96.8 & 32.0 \\
                              & Epoch & 16 & 8 & 8 & 8 & 8 \\
                              & Batch & 4 & 4 & 8 & 4 & 32\\
\bottomrule[2pt]
\end{tabular}
}
\caption{Comparison of VRAM and GPU usage for BitDistiller and RCP.}
\label{tab:cost}
\end{table}
Table~\ref{tab:cost} summarizes the training configurations and training costs.
The VRAM usage denotes the memory consumed on a single GPU and the GPU-hours were calculated by multiplying the training time by the total number of GPUs used. In our experiments, we conducted experiments on LLaMA-1, LLaMA-2 7B and LLaMA-3.2 (1B and 3B) on 8 RTX A6000 GPUs (48 GB each). For larger-scale models, LLaMA-3 8B was trained on 8 A100 GPUs (80GB each), and LLaMA-2 13B was trained on 8 GPUs (141GB each) within a DGX H200 system. The enlarged vocabulary in LLaMA-3 and later models increases gradient-computation demands, resulting in higher VRAM usage. To ensure training stability under these constraints, we set the training batch size to 4. 


Our proposed RCP incurs approximately 10\% additional training cost compared to the baseline. However, on LLaMA-2 7B, it improves perplexity from 17.40 to 8.31-nearly a twofold improvement. Furthermore, for LLaMA-3.2 3B, RCP improves the PPL by up to 42 times compared to the baseline. Considering these significant performance gains, the 10\% additional training cost is acceptable.
\subsection{Ablation Studies}
\begin{table}[htbp]
\centering
\renewcommand{\arraystretch}{0.8}
\resizebox{0.8\columnwidth}{!}{ 
{\small
\begin{tabular}{c|c|c|c|c}
\toprule[1pt]
\textbf{\#Bits} & \textbf{Rotation} & \textbf{LWC} & \textbf{LDP} & \textbf{PPL$^\downarrow$} \\ \hline
\multirow{4}{*}{2-4-4} &  &  &  & 17.40 \\ 
                       & \checkmark &  &  & 8.93 \\ 
                       & \checkmark & \checkmark & \ & 8.79 \\ 
                       & \checkmark & \checkmark & \checkmark & 8.31 \\ \bottomrule[1pt]
\end{tabular}
}
}
\captionsetup{justification=justified, singlelinecheck=false}
\caption{Ablation study on the impact of each component of RCP on performance for LLaMA-2 7B.}
\label{tab:ablation}
\end{table}

\paragraph{Impact of RCP Components}
As shown in Table \ref{tab:ablation}, we conducted an ablation study to analyze the impact of removing each component of RCP on model performance. In 4-bit activation quantization, addressing the outliers in activations was crucial, and this was effectively resolved using rotation, which led to the largest performance gain compared to the baseline. This demonstrates that rotation is a viable solution when quantizing activations to low bit-width.
Learning the clipping range with LWC (Rotation $\checkmark$ and LWC $\checkmark$) gives a small extra reduction in PPL. This shows that the initial range computed by the grid search algorithm of AWQ is already reasonable, but adjusting the clipping parameters to better fit the updated weight can be beneficial for QAT. On top of these two components, LDP further reduces quantization error under extremely low bit-width settings.
\paragraph{Impact of Rotation Configuration}
\begin{table}[htbp]
\centering
\renewcommand{\arraystretch}{0.6}
\resizebox{0.5\columnwidth}{!}{ 
\begin{tabular}{lcc}
\toprule[1pt]
\textbf{W2A4KV4} & \textbf{PPL$^\downarrow$}   \\ \midrule
RCP       & \textbf{8.31}         \\ \midrule
-$R_3$ & 8.48                     \\
-[$R_2$,$R_3$]   & 8.83                      \\
-[$R_3$,$R_4$]     & 12.24                \\ 
-[$R_2$,$R_3$,$R_4$]     & 12.76                 \\ 
-[$R_1$,$R_2$,$R_3$,$R_4$]     & 25.05                \\ 

\bottomrule[1pt]
\end{tabular}
}
\captionsetup{
    justification=justified,
    singlelinecheck=false
}
\caption{Ablation study on the impact of rotation configuration for LLaMA-2 7B.}
\label{tab:ablation_rotation}
\end{table}
Since the rotation requires additional processes before and after inference, we investigated the performance trend by incrementally adding rotation matrices ($R_1$,$R_2$,$R_3$,$R_4$) to different components to find an appropriate balance between accuracy and overhead. The results are presented in Table \ref{tab:ablation_rotation}. 
The table demonstrates that the impact of the rotation was most significant with $R_1$ and $R_4$. Especially, $R_1$, which applies rotation matrix to the input weight and input activation of all modules thereby having the largest impact on quantization performance. Additionally, our analysis revealed that in LLaMA-2 7B, the input to the down projection layer (of the MLP) exhibited a significant number of outliers, which was effectively addressed through $R_4$ online rotation to activation. 
Additional ablation results can be found in Appendix~\ref{sec:additional_ablation}.


\section{Conclusion}
\label{sec:conclusion}
RCP enables weights to be quantized to extreme low-bit precision through learnable non-uniform quantization while harmonizing with rotation to optimize both activations and KV-cache to 4-bit. RCP has achieved the first W2A4KV4 configuration and implemented optimized kernels for inference, facilitating LLM serving even in resource-constrained environments.
\section*{Limitations}

Although our proposed RCP first enables challenging W2A4KV4 quantization of commonly used LLM models, we report key limitations of our work.

First, the online rotation operators ($R_2$ through $R_4$) inevitably introduce additional latency for training and evaluation. Custom CUDA kernels or FlashAttention3~\cite{flashattention3} can minimize such speed-down, however, it might not be a viable option for many edge application scenarios where no hardware support for fast Hadamard transform is available.

Second, RCP requires heavier hyperparameter tuning than BitDistiller since rotation tends to make the model weights more sensitive to the choice of learning rate. This can be prohibitive when a user is under a strict budget limit.

In future work, we could explore applying an optimized rotation matrix that achieves comparable performance to Cayley-optimized rotation matrices used in SpinQuant~\cite{spinquant} while maintaining similar computational costs to the Random Hadamard rotation matrices employed in QuaRot~\cite{quarot}. \\

\section*{Acknowledgments}
This work was supported by Samsung Advanced Institute of Technology, and MX division, Samsung Electronics Co., Ltd., Inter-university Semiconductor Research Center (ISRC) and Institute of Information \& communications Technology Planning \& Evaluation (IITP) grant funded by the Korea government(MSIT) [NO. RS-2021-II211343, Artificial Intelligence Graduate School Program (Seoul National University)].


\bibliography{latex/acl_latex}

\begin{thebibliography}{40}
\providecommand{\natexlab}[1]{#1}

\bibitem[{AI@Meta(2024)}]{llama3modelcard}
AI@Meta. 2024.
\newblock \href {https://github.com/meta-llama/llama3/blob/main/MODEL_CARD.md} {Llama 3 model card}.

\bibitem[{Ashkboos et~al.(2024{\natexlab{a}})Ashkboos, Croci, do~Nascimento, Hoefler, and Hensman}]{slicegpt}
Saleh Ashkboos, Maximilian~L. Croci, Marcelo~Gennari do~Nascimento, Torsten Hoefler, and James Hensman. 2024{\natexlab{a}}.
\newblock \href {https://openreview.net/forum?id=vXxardq6db} {Slice{GPT}: Compress large language models by deleting rows and columns}.
\newblock In \emph{The Twelfth International Conference on Learning Representations}.

\bibitem[{Ashkboos et~al.(2024{\natexlab{b}})Ashkboos, Mohtashami, Croci, Li, Jaggi, Alistarh, Hoefler, and Hensman}]{quarot}
Saleh Ashkboos, Amirkeivan Mohtashami, Maximilian~L Croci, Bo~Li, Martin Jaggi, Dan Alistarh, Torsten Hoefler, and James Hensman. 2024{\natexlab{b}}.
\newblock Quarot: Outlier-free 4-bit inference in rotated llms.
\newblock \emph{arXiv preprint arXiv:2404.00456}.

\bibitem[{Bai et~al.(2024)Bai, Lv, Zhang, Lyu, Tang, Huang, Du, Liu, Zeng, Hou, Dong, Tang, and Li}]{longbench}
Yushi Bai, Xin Lv, Jiajie Zhang, Hongchang Lyu, Jiankai Tang, Zhidian Huang, Zhengxiao Du, Xiao Liu, Aohan Zeng, Lei Hou, Yuxiao Dong, Jie Tang, and Juanzi Li. 2024.
\newblock \href {https://arxiv.org/abs/2308.14508} {Longbench: A bilingual, multitask benchmark for long context understanding}.
\newblock \emph{Preprint}, arXiv:2308.14508.

\bibitem[{Bisk et~al.(2020)Bisk, Zellers, Gao, Choi et~al.}]{piqa}
Yonatan Bisk, Rowan Zellers, Jianfeng Gao, Yejin Choi, et~al. 2020.
\newblock Piqa: Reasoning about physical commonsense in natural language.
\newblock In \emph{Proceedings of the AAAI conference on artificial intelligence}, volume~34, pages 7432--7439.

\bibitem[{Chee et~al.(2024)Chee, Cai, Kuleshov, and Sa}]{quip}
Jerry Chee, Yaohui Cai, Volodymyr Kuleshov, and Christopher~De Sa. 2024.
\newblock \href {https://arxiv.org/abs/2307.13304} {Quip: 2-bit quantization of large language models with guarantees}.
\newblock \emph{Preprint}, arXiv:2307.13304.

\bibitem[{Chen et~al.(2021)Chen, Tworek, Jun, Yuan, Pinto, Kaplan, Edwards, Burda, Joseph, Brockman et~al.}]{humaneval}
Mark Chen, Jerry Tworek, Heewoo Jun, Qiming Yuan, Henrique Ponde De~Oliveira Pinto, Jared Kaplan, Harri Edwards, Yuri Burda, Nicholas Joseph, Greg Brockman, et~al. 2021.
\newblock Evaluating large language models trained on code.
\newblock \emph{arXiv preprint arXiv:2107.03374}.

\bibitem[{Chen et~al.(2024)Chen, Shao, Xu, Wang, Gao, Zhang, Qiao, and Luo}]{efficientqat}
Mengzhao Chen, Wenqi Shao, Peng Xu, Jiahao Wang, Peng Gao, Kaipeng Zhang, Yu~Qiao, and Ping Luo. 2024.
\newblock Efficientqat: Efficient quantization-aware training for large language models.
\newblock \emph{arXiv preprint arXiv:2407.11062}.

\bibitem[{Choi et~al.(2018)Choi, Wang, Venkataramani, Chuang, Srinivasan, and Gopalakrishnan}]{pact}
Jungwook Choi, Zhuo Wang, Swagath Venkataramani, Pierce I-Jen Chuang, Vijayalakshmi Srinivasan, and Kailash Gopalakrishnan. 2018.
\newblock \href {https://arxiv.org/abs/1805.06085} {Pact: Parameterized clipping activation for quantized neural networks}.
\newblock \emph{Preprint}, arXiv:1805.06085.

\bibitem[{Clark et~al.(2018)Clark, Cowhey, Etzioni, Khot, Sabharwal, Schoenick, and Tafjord}]{arc}
Peter Clark, Isaac Cowhey, Oren Etzioni, Tushar Khot, Ashish Sabharwal, Carissa Schoenick, and Oyvind Tafjord. 2018.
\newblock Think you have solved question answering? try arc, the ai2 reasoning challenge.
\newblock \emph{arXiv preprint arXiv:1803.05457}.

\bibitem[{Cobbe et~al.(2021)Cobbe, Kosaraju, Bavarian, Chen, Jun, Kaiser, Plappert, Tworek, Hilton, Nakano, Hesse, and Schulman}]{gsm8k}
Karl Cobbe, Vineet Kosaraju, Mohammad Bavarian, Mark Chen, Heewoo Jun, Lukasz Kaiser, Matthias Plappert, Jerry Tworek, Jacob Hilton, Reiichiro Nakano, Christopher Hesse, and John Schulman. 2021.
\newblock Training verifiers to solve math word problems.
\newblock \emph{arXiv preprint arXiv:2110.14168}.

\bibitem[{Du et~al.(2024)Du, Zhang, Cao, Guo, Cao, Chu, and Xu}]{bitdistiller}
Dayou Du, Yijia Zhang, Shijie Cao, Jiaqi Guo, Ting Cao, Xiaowen Chu, and Ningyi Xu. 2024.
\newblock Bitdistiller: Unleashing the potential of sub-4-bit llms via self-distillation.
\newblock \emph{arXiv preprint arXiv:2402.10631}.

\bibitem[{Esser et~al.(2020)Esser, McKinstry, Bablani, Appuswamy, and Modha}]{lsq}
Steven~K. Esser, Jeffrey~L. McKinstry, Deepika Bablani, Rathinakumar Appuswamy, and Dharmendra~S. Modha. 2020.
\newblock \href {https://openreview.net/forum?id=rkgO66VKDS} {Learned step size quantization}.
\newblock In \emph{International Conference on Learning Representations}.

\bibitem[{Frantar et~al.(2022)Frantar, Ashkboos, Hoefler, and Alistarh}]{gptq}
Elias Frantar, Saleh Ashkboos, Torsten Hoefler, and Dan Alistarh. 2022.
\newblock Optq: Accurate quantization for generative pre-trained transformers.
\newblock In \emph{The Eleventh International Conference on Learning Representations}.

\bibitem[{Guo et~al.(2024)Guo, Brandon, Cholakov, Ragan-Kelley, Xing, and Kim}]{flute}
Han Guo, William Brandon, Radostin Cholakov, Jonathan Ragan-Kelley, Eric~P. Xing, and Yoon Kim. 2024.
\newblock \href {https://arxiv.org/abs/2407.10960} {Fast matrix multiplications for lookup table-quantized llms}.
\newblock \emph{Preprint}, arXiv:2407.10960.

\bibitem[{Hendrycks et~al.(2020)Hendrycks, Burns, Basart, Zou, Mazeika, Song, and Steinhardt}]{mmlu}
Dan Hendrycks, Collin Burns, Steven Basart, Andy Zou, Mantas Mazeika, Dawn Song, and Jacob Steinhardt. 2020.
\newblock Measuring massive multitask language understanding.
\newblock \emph{arXiv preprint arXiv:2009.03300}.

\bibitem[{Hinton et~al.(2015)Hinton, Vinyals, and Dean}]{kd}
Geoffrey Hinton, Oriol Vinyals, and Jeff Dean. 2015.
\newblock Distilling the knowledge in a neural network.
\newblock \emph{arXiv preprint arXiv:1503.02531}.

\bibitem[{Kim et~al.(2024)Kim, Hooper, Gholami, Dong, Li, Shen, Mahoney, and Keutzer}]{squeezellm}
Sehoon Kim, Coleman Richard~Charles Hooper, Amir Gholami, Zhen Dong, Xiuyu Li, Sheng Shen, Michael~W. Mahoney, and Kurt Keutzer. 2024.
\newblock \href {https://openreview.net/forum?id=0jpbpFia8m} {Squeeze{LLM}: Dense-and-sparse quantization}.
\newblock In \emph{Forty-first International Conference on Machine Learning}.

\bibitem[{Lin et~al.(2024{\natexlab{a}})Lin, Xu, Wu, Cui, Zhang, Mou, Song, Sun, and Wei}]{duquant}
Haokun Lin, Haobo Xu, Yichen Wu, Jingzhi Cui, Yingtao Zhang, Linzhan Mou, Linqi Song, Zhenan Sun, and Ying Wei. 2024{\natexlab{a}}.
\newblock \href {https://arxiv.org/abs/2406.01721} {Duquant: Distributing outliers via dual transformation makes stronger quantized llms}.
\newblock \emph{Preprint}, arXiv:2406.01721.

\bibitem[{Lin et~al.(2024{\natexlab{b}})Lin, Tang, Tang, Yang, Chen, Wang, Xiao, Dang, Gan, and Han}]{awq}
Ji~Lin, Jiaming Tang, Haotian Tang, Shang Yang, Wei-Ming Chen, Wei-Chen Wang, Guangxuan Xiao, Xingyu Dang, Chuang Gan, and Song Han. 2024{\natexlab{b}}.
\newblock Awq: Activation-aware weight quantization for llm compression and acceleration.
\newblock In \emph{MLSys}.

\bibitem[{Lin et~al.(2024{\natexlab{c}})Lin, Tang, Yang, Zhang, Xiao, Gan, and Han}]{qserve}
Yujun Lin, Haotian Tang, Shang Yang, Zhekai Zhang, Guangxuan Xiao, Chuang Gan, and Song Han. 2024{\natexlab{c}}.
\newblock \href {https://arxiv.org/abs/2405.04532} {Qserve: W4a8kv4 quantization and system co-design for efficient llm serving}.
\newblock \emph{Preprint}, arXiv:2405.04532.

\bibitem[{Liu et~al.(2022)Liu, Cheng, Huang, Xing, and Shen}]{nu2u}
Zechun Liu, Kwang-Ting Cheng, Dong Huang, Eric~P Xing, and Zhiqiang Shen. 2022.
\newblock Nonuniform-to-uniform quantization: Towards accurate quantization via generalized straight-through estimation.
\newblock In \emph{Proceedings of the IEEE/CVF conference on computer vision and pattern recognition}, pages 4942--4952.

\bibitem[{Liu et~al.(2024{\natexlab{a}})Liu, Oguz, Zhao, Chang, Stock, Mehdad, Shi, Krishnamoorthi, and Chandra}]{llmqat}
Zechun Liu, Barlas Oguz, Changsheng Zhao, Ernie Chang, Pierre Stock, Yashar Mehdad, Yangyang Shi, Raghuraman Krishnamoorthi, and Vikas Chandra. 2024{\natexlab{a}}.
\newblock Llm-qat: Data-free quantization aware training for large language models.
\newblock \emph{arXiv preprint arXiv:2305.17888}.

\bibitem[{Liu et~al.(2024{\natexlab{b}})Liu, Zhao, Fedorov, Soran, Choudhary, Krishnamoorthi, Chandra, Tian, and Blankevoort}]{spinquant}
Zechun Liu, Changsheng Zhao, Igor Fedorov, Bilge Soran, Dhruv Choudhary, Raghuraman Krishnamoorthi, Vikas Chandra, Yuandong Tian, and Tijmen Blankevoort. 2024{\natexlab{b}}.
\newblock Spinquant--llm quantization with learned rotations.
\newblock \emph{arXiv preprint arXiv:2405.16406}.

\bibitem[{Luo et~al.(2023)Luo, Xu, Zhao, Sun, Geng, Hu, Tao, Ma, Lin, and Jiang}]{luo2023wizardcoder}
Ziyang Luo, Can Xu, Pu~Zhao, Qingfeng Sun, Xiubo Geng, Wenxiang Hu, Chongyang Tao, Jing Ma, Qingwei Lin, and Daxin Jiang. 2023.
\newblock Wizardcoder: Empowering code large language models with evol-instruct.
\newblock \emph{arXiv preprint arXiv:2306.08568}.

\bibitem[{Merity et~al.(2016)Merity, Xiong, Bradbury, and Socher}]{wiki2}
Stephen Merity, Caiming Xiong, James Bradbury, and Richard Socher. 2016.
\newblock Pointer sentinel mixture models.
\newblock \emph{arXiv preprint arXiv:1609.07843}.

\bibitem[{Paszke et~al.(2019)Paszke, Gross, Massa, Lerer, Bradbury, Chanan, Killeen, Lin, Gimelshein, Antiga et~al.}]{pytorch}
Adam Paszke, Sam Gross, Francisco Massa, Adam Lerer, James Bradbury, Gregory Chanan, Trevor Killeen, Zeming Lin, Natalia Gimelshein, Luca Antiga, et~al. 2019.
\newblock Pytorch: An imperative style, high-performance deep learning library.
\newblock \emph{Advances in neural information processing systems}, 32.

\bibitem[{Sakaguchi et~al.(2021)Sakaguchi, Bras, Bhagavatula, and Choi}]{winogrande}
Keisuke Sakaguchi, Ronan~Le Bras, Chandra Bhagavatula, and Yejin Choi. 2021.
\newblock Winogrande: An adversarial winograd schema challenge at scale.
\newblock \emph{Communications of the ACM}, 64(9):99--106.

\bibitem[{Shah et~al.(2024)Shah, Bikshandi, Zhang, Thakkar, Ramani, and Dao}]{flashattention3}
Jay Shah, Ganesh Bikshandi, Ying Zhang, Vijay Thakkar, Pradeep Ramani, and Tri Dao. 2024.
\newblock Flashattention-3: Fast and accurate attention with asynchrony and low-precision.
\newblock \emph{arXiv preprint arXiv:2407.08608}.

\bibitem[{Shao et~al.(2024)Shao, Chen, Zhang, Xu, Zhao, Li, Zhang, Gao, Qiao, and Luo}]{omniquant}
Wenqi Shao, Mengzhao Chen, Zhaoyang Zhang, Peng Xu, Lirui Zhao, Zhiqian Li, Kaipeng Zhang, Peng Gao, Yu~Qiao, and Ping Luo. 2024.
\newblock \href {https://openreview.net/forum?id=8Wuvhh0LYW} {Omniquant: Omnidirectionally calibrated quantization for large language models}.
\newblock In \emph{The Twelfth International Conference on Learning Representations}.

\bibitem[{Taori et~al.(2023)Taori, Gulrajani, Zhang, Dubois, Li, Guestrin, Liang, and Hashimoto}]{alpaca}
Rohan Taori, Ishaan Gulrajani, Tianyi Zhang, Yann Dubois, Xuechen Li, Carlos Guestrin, Percy Liang, and Tatsunori~B. Hashimoto. 2023.
\newblock Stanford alpaca: An instruction-following llama model.
\newblock \url{https://github.com/tatsu-lab/stanford_alpaca}.

\bibitem[{Touvron et~al.(2023{\natexlab{a}})Touvron, Lavril, Izacard, Martinet, Lachaux, Lacroix, Rozi{\`e}re, Goyal, Hambro, Azhar et~al.}]{llama}
Hugo Touvron, Thibaut Lavril, Gautier Izacard, Xavier Martinet, Marie-Anne Lachaux, Timoth{\'e}e Lacroix, Baptiste Rozi{\`e}re, Naman Goyal, Eric Hambro, Faisal Azhar, et~al. 2023{\natexlab{a}}.
\newblock Llama: Open and efficient foundation language models.
\newblock \emph{arXiv preprint arXiv:2302.13971}.

\bibitem[{Touvron et~al.(2023{\natexlab{b}})Touvron, Martin, Stone, Albert, Almahairi, Babaei, Bashlykov, Batra, Bhargava, Bhosale et~al.}]{llama2}
Hugo Touvron, Louis Martin, Kevin Stone, Peter Albert, Amjad Almahairi, Yasmine Babaei, Nikolay Bashlykov, Soumya Batra, Prajjwal Bhargava, Shruti Bhosale, et~al. 2023{\natexlab{b}}.
\newblock Llama 2: Open foundation and fine-tuned chat models.
\newblock \emph{arXiv preprint arXiv:2307.09288}.

\bibitem[{Tseng et~al.(2024)Tseng, Chee, Sun, Kuleshov, and Sa}]{quip-sharp}
Albert Tseng, Jerry Chee, Qingyao Sun, Volodymyr Kuleshov, and Christopher~De Sa. 2024.
\newblock \href {https://arxiv.org/abs/2402.04396} {Quip\#: Even better llm quantization with hadamard incoherence and lattice codebooks}.
\newblock \emph{Preprint}, arXiv:2402.04396.

\bibitem[{Wang et~al.(2022)Wang, Dong, Wang, Liu, An, and Guo}]{llt}
Longguang Wang, Xiaoyu Dong, Yingqian Wang, Li~Liu, Wei An, and Yulan Guo. 2022.
\newblock Learnable lookup table for neural network quantization.
\newblock In \emph{Proceedings of the IEEE/CVF conference on computer vision and pattern recognition}, pages 12423--12433.

\bibitem[{Xiao et~al.(2023)Xiao, Lin, Seznec, Wu, Demouth, and Han}]{smoothquant}
Guangxuan Xiao, Ji~Lin, Mickael Seznec, Hao Wu, Julien Demouth, and Song Han. 2023.
\newblock Smoothquant: Accurate and efficient post-training quantization for large language models.
\newblock In \emph{Proceedings of the 40th International Conference on Machine Learning}.

\bibitem[{Yu et~al.(2023)Yu, Jiang, Shi, Yu, Liu, Zhang, Kwok, Li, Weller, and Liu}]{metamathQA}
Longhui Yu, Weisen Jiang, Han Shi, Jincheng Yu, Zhengying Liu, Yu~Zhang, James~T Kwok, Zhenguo Li, Adrian Weller, and Weiyang Liu. 2023.
\newblock Metamath: Bootstrap your own mathematical questions for large language models.
\newblock \emph{arXiv preprint arXiv:2309.12284}.

\bibitem[{Zellers et~al.(2019)Zellers, Holtzman, Bisk, Farhadi, and Choi}]{hellaswag}
Rowan Zellers, Ari Holtzman, Yonatan Bisk, Ali Farhadi, and Yejin Choi. 2019.
\newblock Hellaswag: Can a machine really finish your sentence?
\newblock \emph{arXiv preprint arXiv:1905.07830}.

\bibitem[{Zhang et~al.(2023)Zhang, Zhang, Cao, Du, Wei, Cao, and Xu}]{afpq}
Yijia Zhang, Sicheng Zhang, Shijie Cao, Dayou Du, Jianyu Wei, Ting Cao, and Ningyi Xu. 2023.
\newblock Afpq: Asymmetric floating point quantization for llms.
\newblock \emph{arXiv preprint arXiv:2311.01792}.

\bibitem[{Zhao et~al.(2024)Zhao, Lin, Zhu, Ye, Chen, Zheng, Ceze, Krishnamurthy, Chen, and Kasikci}]{atom}
Yilong Zhao, Chien-Yu Lin, Kan Zhu, Zihao Ye, Lequn Chen, Size Zheng, Luis Ceze, Arvind Krishnamurthy, Tianqi Chen, and Baris Kasikci. 2024.
\newblock \href {https://arxiv.org/abs/2310.19102} {Atom: Low-bit quantization for efficient and accurate llm serving}.
\newblock \emph{Preprint}, arXiv:2310.19102.

\end{thebibliography}

\appendix

\section{Appendix}
\label{sec:appendix}

\subsection{Related Works}
\label{sec:related}
\paragraph{PTQ and QAT}
GPTQ~\cite{gptq} introduced an accurate post-training quantization (PTQ) method based on approximate second-order information that enables weight-only quantization down to 3-4 bits through block-wise reconstruction.
SmoothQuant~\cite{smoothquant} proposed smoothing activation outliers by offline migrating quantization difficulty from activations to weights through equivalent transformation, enabling accurate 8-bit weight-activation quantization.
AWQ~\cite{awq} built upon SmoothQuant's equivalent transformation concept but introduced activation-aware channel-wise scaling to protect salient weights during weight-only quantization. 
OmniQuant~\cite{omniquant} enhanced quantization by introducing learnable weight clipping and equivalent transformation parameters that are jointly optimized through block-wise reconstruction.\\
LLM-QAT~\cite{llmqat} was the first to explore quantization-aware training (QAT) for LLMs using data-free knowledge distillation from the full-precision model to guide low-bit quantization.
BitDistiller~\cite{bitdistiller} improved upon LLM-QAT by introducing a self-distillation framework with confidence-aware KL divergence to enable sub-4-bit quantization while maintaining efficiency.
EfficientQAT~\cite{efficientqat} made QAT more practical by introducing block-wise training of all parameters followed by end-to-end training of quantization parameters.
\paragraph{Rotation}
QuaRot~\cite{quarot} introduced a rotation-based approach using Hadamard transforms to eliminate outliers in activations and KV-cache, enabling end-to-end 4-bit quantization including weights, activations and KV-cache.
SpinQuant~\cite{spinquant} enhanced this rotation-based approach by learning optimal rotation matrices instead of using random ones. 
\paragraph{Non-uniform Quantization}
PACT~\cite{pact} introduced a learnable clipping parameter for activation quantization during training to help preserve model accuracy. SqueezeLLM~\cite{squeezellm} took a different direction by focusing on identifying and extracting outlier values into a sparse format while quantizing the remaining dense values. NU2U~\cite{nu2u} proposed learning flexible non-uniform input thresholds while maintaining uniform output levels to balance quantization accuracy with hardware efficiency.
\paragraph{Serving Optimization}
Atom~\cite{atom} first introduced W4A4 quantization for LLM serving but faced performance challenges from dequantization overhead. QServe~\cite{qserve} addressed the challenges by introducing W4A8KV4 quantization with progressive group quantization FLUTE~\cite{flute} focused on developing efficient GPU kernels for flexible lookup table-based quantization methods that can support arbitrary bit widths including 3-bit and 4-bit quantization.

\subsection{Proof of Lemma 1}\label{sec:lemma1_proof}

\begin{proof}[Proof of Lemma~\ref{lemma:hadamard_kurtosis}]
The proof follows directly from Sub-lemmas~\ref{sublemma:representation}--\ref{sublemma:application}. Specifically, Sublemma~\ref{sublemma:application} shows that for all components $Y_i$ of the transformed vector:
\begin{equation*}
\text{Kurt}(Y_i) = \frac{\text{Kurt}(X)}{n}
\end{equation*}

Since $\text{Kurt}(X) < 0$ for platykurtic distributions and $n > 1$:
\begin{equation*}
\frac{\text{Kurt}(X)}{n} > \text{Kurt}(X)
\end{equation*}

Therefore:
\begin{equation*}
\text{Kurt}(Y_i) > \text{Kurt}(X)
\end{equation*}
for all $i \in \{1, 2, \ldots, n\}$, which completes the proof.
\end{proof}

\begin{definition*}
\label{def:normalized_hadamard}
The $n \times n$ normalized Hadamard matrix $\mathbf{H}_n$, where $n = 2^k$ for some non-negative integer $k$, is defined as:
\begin{equation*}
\mathbf{H}_n = \frac{1}{\sqrt{n}} \mathbf{H}'_n
\end{equation*}
where $\mathbf{H}'_n$ is the unnormalized Hadamard matrix with elements $H'_{ij} \in \{-1, 1\}$ constructed recursively as:
\begin{equation*}
\mathbf{H}'_1 = [1], \quad \mathbf{H}'_{2^{k+1}} = 
\begin{bmatrix} 
\mathbf{H}'_{2^k} & \mathbf{H}'_{2^k} \\ 
\mathbf{H}'_{2^k} & -\mathbf{H}'_{2^k} 
\end{bmatrix}
\end{equation*}
Note that the first row of $\mathbf{H}'_n$ consists entirely of 1s, while every other row contains exactly $n/2$ entries of 1 and $n/2$ entries of $-1$.

Furthermore, the normalized Hadamard matrix $\mathbf{H}_n$ is orthogonal:
\begin{equation*}
\mathbf{H}_n \mathbf{H}_n^T = \mathbf{I}_n
\end{equation*}
where $\mathbf{I}_n$ is the $n \times n$ identity matrix. 
For any random vector $\mathbf{X}$ with independent components of identical variance $\sigma^2$, the transformed vector $\mathbf{Y} = \mathbf{H}_n\mathbf{X}$ has the same component-wise variance:
\begin{equation}
\text{Var}(Y_i) = \sigma^2 \quad \text{for all } i \in \{1, 2, \ldots, n\}
\end{equation}
This follows from the fact that for a covariance matrix $\Sigma_X = \sigma^2 \mathbf{I}_n$, the transformed covariance is $\Sigma_Y = \mathbf{H}_n \Sigma_X \mathbf{H}_n^T = \sigma^2 \mathbf{H}_n \mathbf{H}_n^T = \sigma^2 \mathbf{I}_n = \Sigma_X$.
\end{definition*}

\begin{sublemma}
\label{sublemma:representation}
Each component $Y_i$ of the transformed vector can be expressed as a linear combination of the original variables:
\begin{equation}
Y_i = \frac{1}{\sqrt{n}}\sum_{j=1}^{n} H'_{ij}X_j
\end{equation}
where $H'_{ij} \in \{-1, 1\}$ are the elements of the unnormalized Hadamard matrix.
\end{sublemma}

\begin{proof}
By definition of matrix multiplication, each component $Y_i$ of the transformed vector $\mathbf{Y} = \mathbf{H}_n\mathbf{X}$ is given by:
\begin{equation*}
Y_i = \sum_{j=1}^{n} h_{ij}X_j = \frac{1}{\sqrt{n}}\sum_{j=1}^{n} H'_{ij}X_j
\end{equation*}
where $h_{ij} = \frac{H'_{ij}}{\sqrt{n}}$ are the elements of the normalized Hadamard matrix.
\end{proof}

\begin{sublemma}
\label{sublemma:properties}
If $X_i$ has mean $\mu$ and variance $\sigma^2$, then $Y_i$ has mean $\mu_Y$ and variance $\sigma_Y^2$ where:
\begin{equation*}
\mu_Y = \begin{cases} 
\sqrt{n}\mu & \text{if } i = 1 \\
0 & \text{if } i \neq 1
\end{cases}
\end{equation*}
\begin{equation*}
\sigma_Y^2 = \sigma^2
\end{equation*}
\end{sublemma}

\begin{proof}
Let's calculate the mean of each transformed component $Y_i$:
\begin{equation*}
\begin{split}
\mathbb{E}[Y_i] = \mathbb{E}\left[\frac{1}{\sqrt{n}}\sum_{j=1}^{n} H'_{ij}X_j\right] \\
= \frac{1}{\sqrt{n}}\sum_{j=1}^{n} H'_{ij}\mathbb{E}[X_j] = \frac{\mu}{\sqrt{n}}\sum_{j=1}^{n} H'_{ij}
\end{split}
\end{equation*}

For $i = 1$, the first row of the unnormalized Hadamard matrix consists entirely of 1s. Therefore:
\begin{equation*}
\mathbb{E}[Y_1] = \frac{\mu}{\sqrt{n}} \cdot n = \sqrt{n}\mu
\end{equation*}

For all other rows $i > 1$, the Hadamard matrix has the property that each row contains exactly $\frac{n}{2}$ entries of 1 and $\frac{n}{2}$ entries of -1. This gives:
\begin{equation*}
\mathbb{E}[Y_i] = \frac{\mu}{\sqrt{n}}\left(\frac{n}{2} - \frac{n}{2}\right) = 0 \quad \text{for } i > 1
\end{equation*}

For the variance, assuming independence of $X_j$:
\begin{equation*}
\begin{split}
\text{Var}(Y_i) = \text{Var}\left(\frac{1}{\sqrt{n}}\sum_{j=1}^{n} H'_{ij}X_j\right)\\
= \frac{1}{n}\sum_{j=1}^{n} (H'_{ij})^2 \cdot \text{Var}(X_j)
\end{split}
\end{equation*}

Since $(H'_{ij})^2 = 1$ for all $i,j$ and all $X_j$ have variance $\sigma^2$:
\begin{equation*}
\text{Var}(Y_i) = \frac{\sigma^2}{n} \cdot n = \sigma^2 \quad \text{for all } i
\end{equation*}
\end{proof}

\begin{sublemma}[Relationship Between Fourth Moments]
\label{sublemma:fourthmoments}
For a sum of independent random variables with identical distributions, the standardized fourth cumulant (excess kurtosis) of the sum relates to the individual excess kurtosis by:
\begin{equation*}
\text{Kurt}\left(\frac{1}{\sqrt{n}}\sum_{j=1}^{n} \epsilon_j X_j\right) = \frac{\text{Kurt}(X)}{n}
\end{equation*}
where $\epsilon_j \in \{-1, 1\}$ and $X_j$ are i.i.d. with the same distribution as $X$.
\end{sublemma}

\begin{proof}
Let $Z = \frac{1}{\sqrt{n}}\sum_{j=1}^{n} \epsilon_j X_j$ where $\epsilon_j \in \{-1, 1\}$ and $X_j$ are i.i.d. with the same distribution as $X$.

The excess kurtosis of a random variable $W$ is defined as:
\begin{equation*}
\text{Kurt}(W) = \frac{\mathbb{E}[(W-\mathbb{E}[W])^4]}{(\text{Var}(W))^2} - 3
\end{equation*}

For independent random variables, the cumulants of a sum equal the sum of the cumulants. The fourth cumulant $\kappa_4$ corresponds to:
\begin{equation*}
\begin{split}
\kappa_4(W) = \mathbb{E}[(W-\mathbb{E}[W])^4] - 3(\mathbb{E}[(W-\mathbb{E}[W])^2])^2\\
= \text{Var}(W)^2 \cdot \text{Kurt}(W)
\end{split}
\end{equation*}

For our sum $Z$, the fourth cumulant is:
\begin{equation*}
\kappa_4(Z) = \sum_{j=1}^{n} \kappa_4\left(\frac{\epsilon_j X_j}{\sqrt{n}}\right)
\end{equation*}

Since $\kappa_4(\alpha X) = \alpha^4 \kappa_4(X)$ for any scalar $\alpha$:
\begin{equation*}
\kappa_4(Z) = \sum_{j=1}^{n} \frac{\epsilon_j^4}{n^2} \kappa_4(X_j) = \frac{1}{n^2} \sum_{j=1}^{n} \kappa_4(X_j)
\end{equation*}

Given that $\epsilon_j^4 = 1$ and all $X_j$ have the same distribution:
\begin{equation*}
\kappa_4(Z) = \frac{n}{n^2} \kappa_4(X) = \frac{\kappa_4(X)}{n}
\end{equation*}

Since $\kappa_4(X) = \text{Var}(X)^2 \cdot \text{Kurt}(X)$ and $\text{Var}(Z) = \text{Var}(X)$ (as shown in Sublemma~\ref{sublemma:properties}):
\begin{equation*}
\text{Var}(Z)^2 \cdot \text{Kurt}(Z) = \frac{\text{Var}(X)^2 \cdot \text{Kurt}(X)}{n}
\end{equation*}

Therefore:
\begin{equation*}
\text{Kurt}(Z) = \frac{\text{Kurt}(X)}{n}
\end{equation*}
\end{proof}

\begin{sublemma}[Application to Hadamard Transform]
\label{sublemma:application}
For a random vector with i.i.d. components having negative excess kurtosis ($\text{Kurt}(X) < 0$), after applying the normalized Hadamard transform:
\begin{equation*}
\text{Kurt}(Y_i) = \frac{\text{Kurt}(X)}{n}
\end{equation*}
Since $\text{Kurt}(X) < 0$ and $n > 1$, we have $\frac{\text{Kurt}(X)}{n} > \text{Kurt}(X)$, which implies $\text{Kurt}(Y_i) > \text{Kurt}(X)$.
\end{sublemma}

\begin{proof}
From Sublemma~\ref{sublemma:representation}, each component $Y_i$ of the Hadamard transform can be written as:
\begin{equation*}
Y_i = \frac{1}{\sqrt{n}}\sum_{j=1}^{n} H'_{ij}X_j
\end{equation*}

This is precisely the form analyzed in Sublemma~\ref{sublemma:fourthmoments}, with $\epsilon_j = H'_{ij}$.

For $i > 1$ (where the mean is 0), applying Sublemma~\ref{sublemma:fourthmoments} directly:
\begin{equation*}
\text{Kurt}(Y_i) = \frac{\text{Kurt}(X)}{n}
\end{equation*}

For $i = 1$, we need to account for the non-zero mean. We can center the variable:
\begin{equation*}
Y_1 - \mathbb{E}[Y_1] = \frac{1}{\sqrt{n}}\sum_{j=1}^{n} (X_j - \mu)
\end{equation*}

Applying the same cumulant analysis to this centered variable:
\begin{equation*}
\text{Kurt}(Y_1) = \frac{\text{Kurt}(X)}{n}
\end{equation*}

Since we assumed $\text{Kurt}(X) < 0$ for a platykurtic distribution, and $n > 1$:
\begin{equation*}
\frac{\text{Kurt}(X)}{n} > \text{Kurt}(X)
\end{equation*}

Therefore, for all components $i \in \{1, 2, \ldots, n\}$:
\begin{equation*}
\text{Kurt}(Y_i) > \text{Kurt}(X)
\end{equation*}
\end{proof}








\subsection{Additional Results on the Kurtosis Analysis} \label{sec:qerr_vs_kurt_more_results}

\paragraph{Details on Fig. \ref{fig:qerr_vs_kurt}}
Let $T$, $C$, and $H$ denote the sequence length of an input, the input channel of weight, and the output channel of weight, respectively. Since RCP works on group-wise quantization, we additionally denote the group size as $G$ and the number of groups as $N$ so that $C=NG$.
The input activation $\mathbf{X}$ and its rotated version $\mathbf{X}_r$ both have a dimension of $(T, C)$, and the weight $\mathbf{W}$ and its rotated \& clipped version $\mathbf{W}_{rc}$ a dimension of $(C, H)$.

We compute the group-wise excess kurtosis $Kurt_{group}$ by reshaping the weights into $(N, G, H)$, and computing the excess kurtosis along the second dimension, resulting in a shape $(N, H)$. For the mean absolute error $QErr(\mathbf{W}, \mathbf{X})$, the shape of the output activation $\mathbf{XW}$ and the quantized version $\mathbf{X}Q(\mathbf{W})$ is $(T, H)$. Then, we average $Kurt_{group}$ along the $N$ dimension so that the excess kurtosis values of the quantization groups contributing to a single output activation element are averaged. The mean absolute error of the output activation is averaged along the $T$ dimension to measure the mean increase of the quantization error in each output activation element.

In all two-dimensional histogram plots, the range is limited to $[-1.5\sigma,1.5\sigma]$ for both axes to prevent outliers from occupying most of the space.

\begin{figure*}
    \centering
    \begin{subfigure}[b]{0.3\textwidth}
        \includegraphics[width=\textwidth]{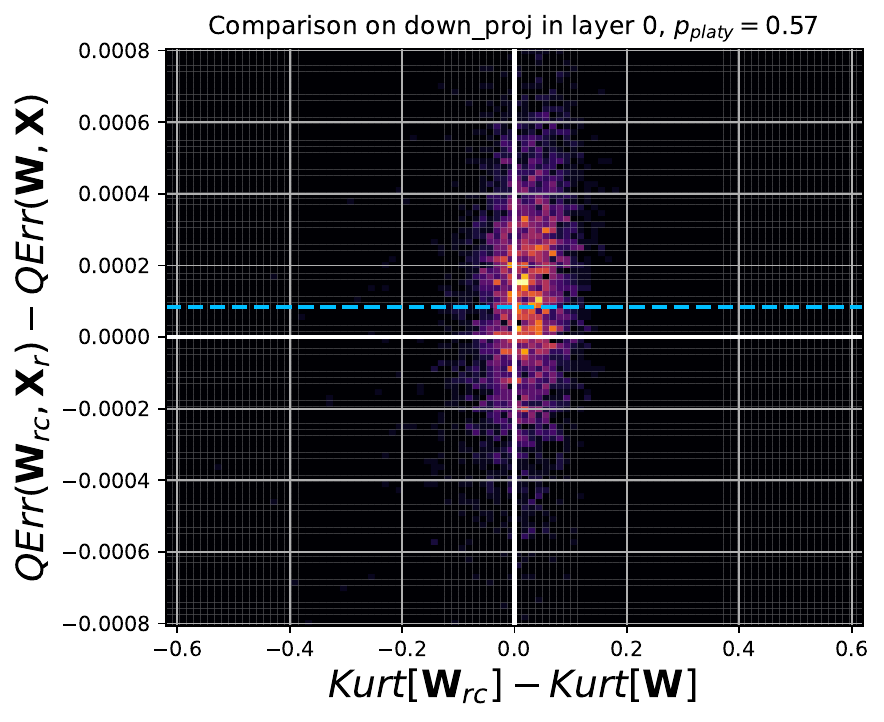}
        \caption{}
        \label{fig:00}
    \end{subfigure}
    \hfill
    \begin{subfigure}[b]{0.3\textwidth}
        \includegraphics[width=\textwidth]{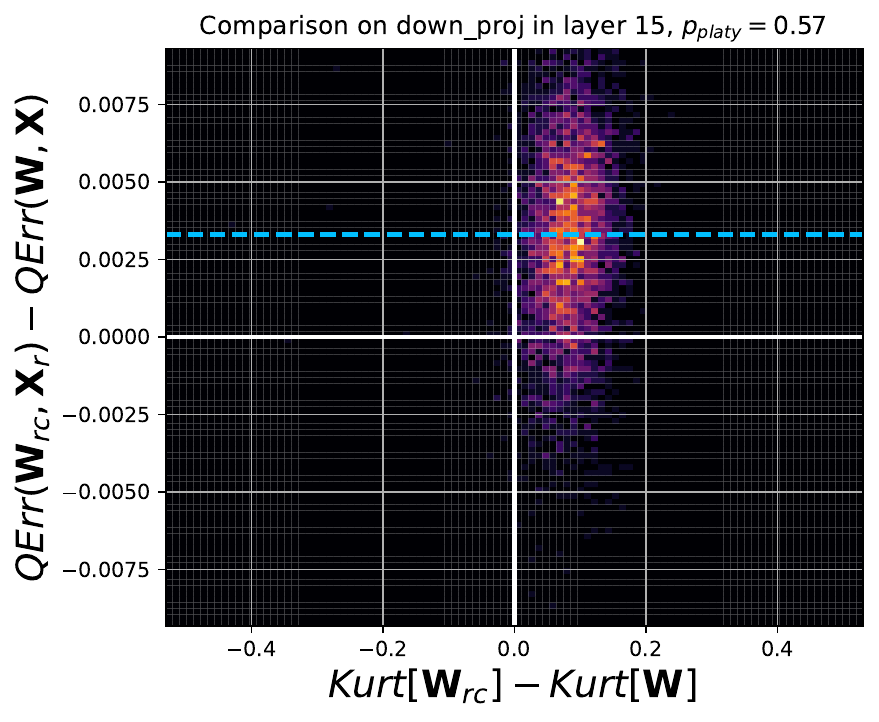}
        \caption{}
        \label{fig:01}
    \end{subfigure}
    \hfill
    \begin{subfigure}[b]{0.3\textwidth}
        \includegraphics[width=\textwidth]{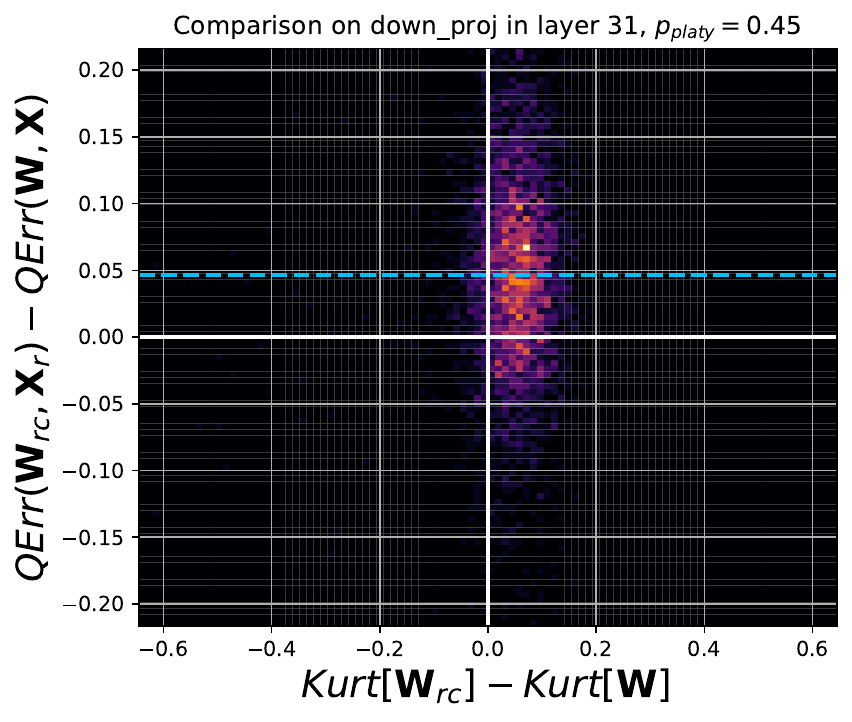}
        \caption{}
        \label{fig:02}
    \end{subfigure}
    
    \begin{subfigure}[b]{0.3\textwidth}
        \includegraphics[width=\textwidth]{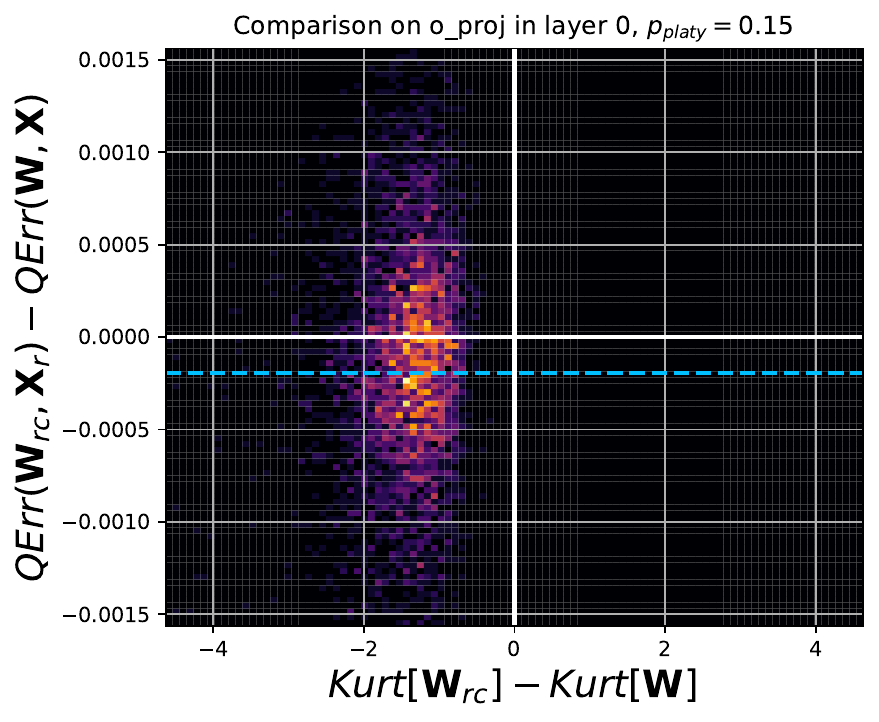}
        \caption{}
        \label{fig:10}
    \end{subfigure}
    \hfill
    \begin{subfigure}[b]{0.3\textwidth}
        \includegraphics[width=\textwidth]{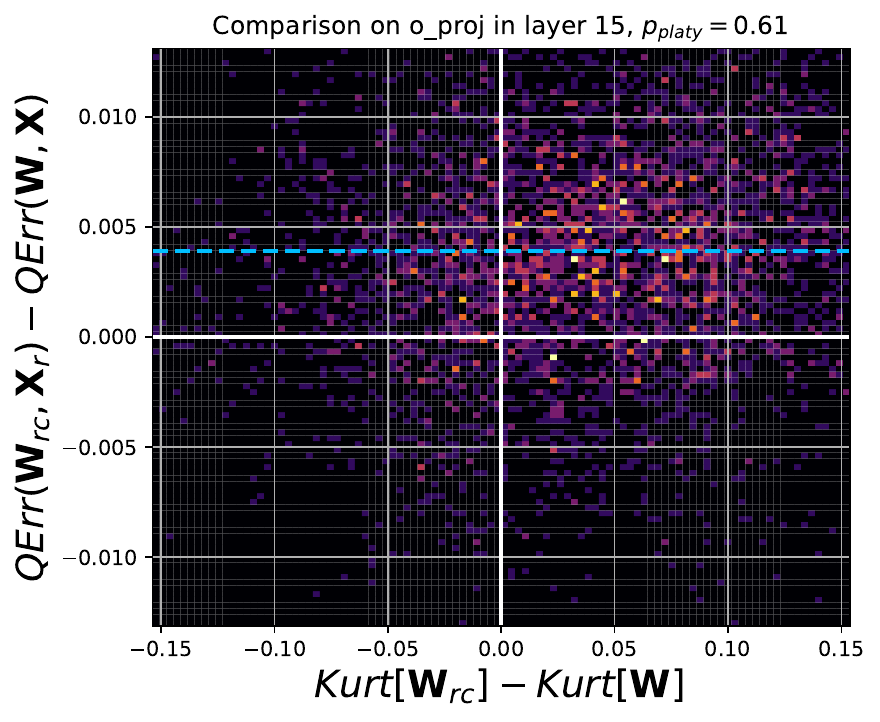}
        \caption{}
        \label{fig:11}
    \end{subfigure}
    \hfill
    \begin{subfigure}[b]{0.3\textwidth}
        \includegraphics[width=\textwidth]{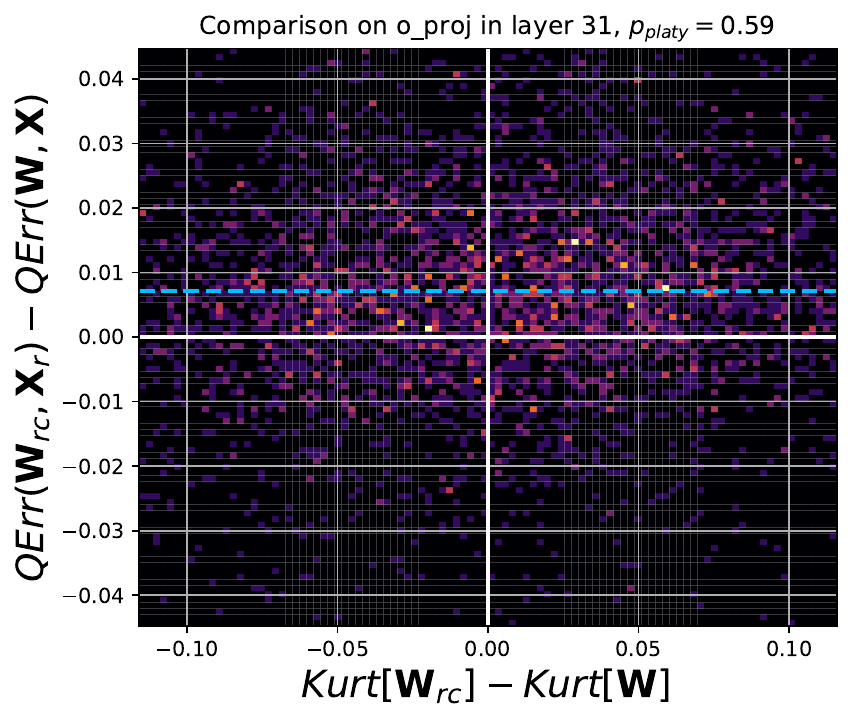}
        \caption{}
        \label{fig:12}
    \end{subfigure}
    
    \begin{subfigure}[b]{0.3\textwidth}
        \includegraphics[width=\textwidth]{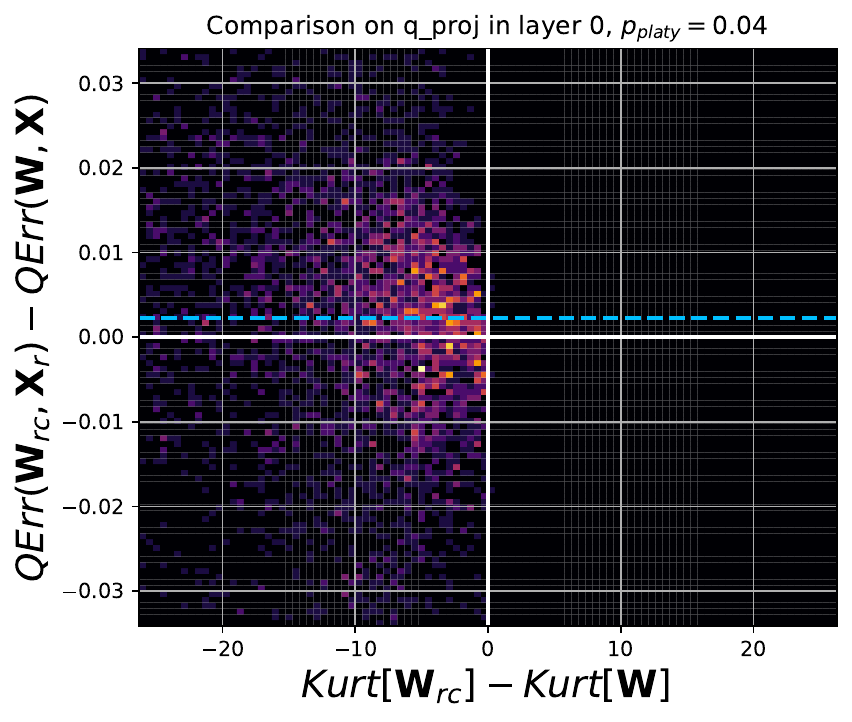}
        \caption{}
        \label{fig:20}
    \end{subfigure}
    \hfill
    \begin{subfigure}[b]{0.3\textwidth}
        \includegraphics[width=\textwidth]{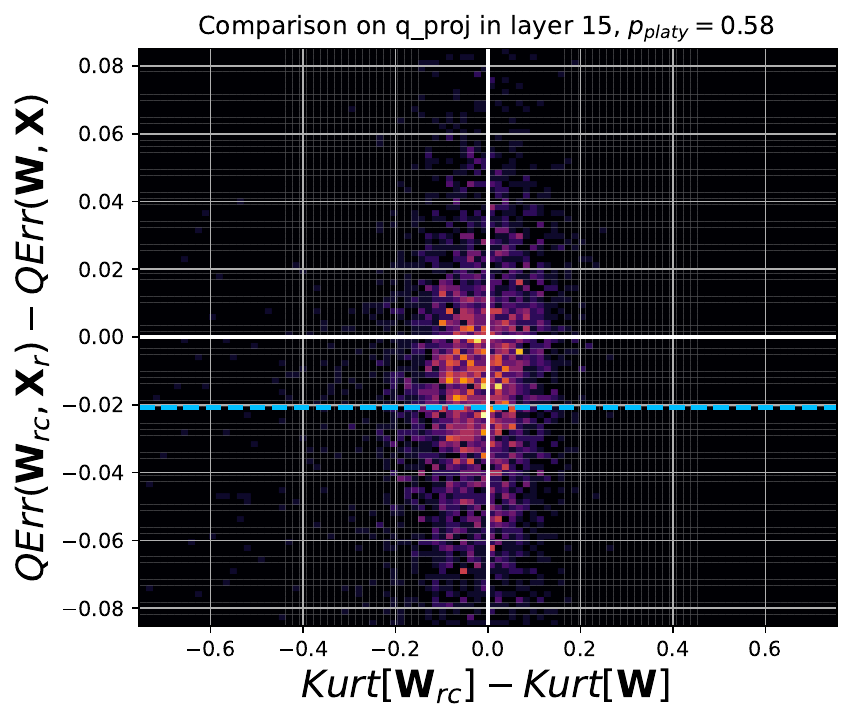}
        \caption{}
        \label{fig:21}
    \end{subfigure}
    \hfill
    \begin{subfigure}[b]{0.3\textwidth}
        \includegraphics[width=\textwidth]{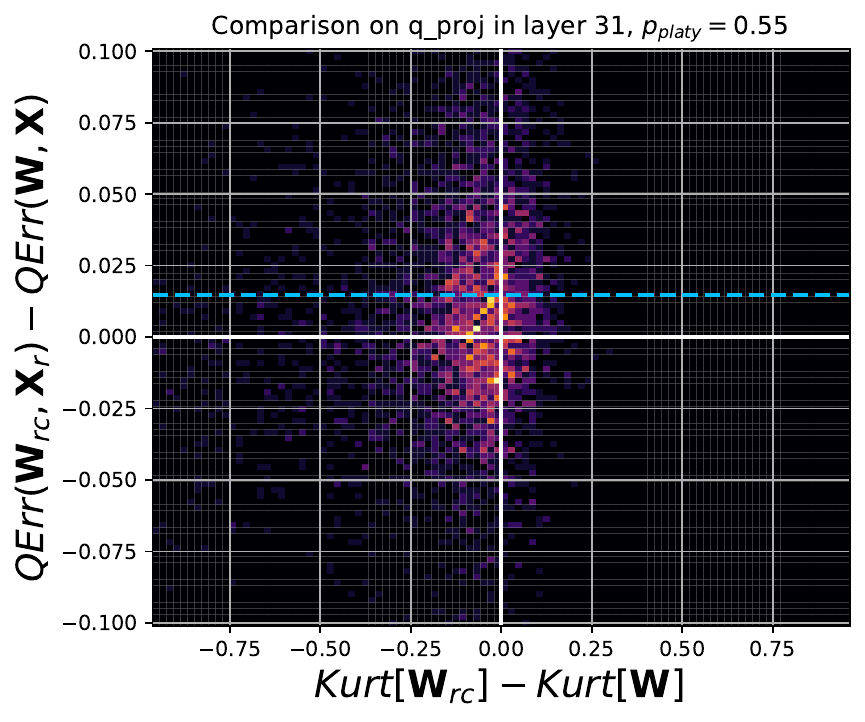}
        \caption{}
        \label{fig:22}
    \end{subfigure}
    
    \caption{Additional two-dimensional histogram plots with $p_{platy}=P(Kurt(\mathbf{W})<0)$ specified in each title.}
    \label{fig:more_qerr_vs_kurt}
\end{figure*}

\paragraph{More Plots and Discussion}
We repeat the same experiments as in Section \ref{sec:kurt_analysis} on three different transformer layers (0, 15, and 31) and three different types of weights ($q\_proj$ , $o\_proj$, and $down\_proj$) and the plots are presented in Fig. \ref{fig:more_qerr_vs_kurt}. In each subplot's title, we specify the ratio of quantization groups that are platykurtic (i.e., $p_{platy}=Kurt(\mathbf{W}_{group})<0$. When $p_{platy} \ll 0.5$ (Fig. \ref{fig:10} and \ref{fig:20}), the Hadamard transform decreases the excess kurtosis, possibly reducing the average quantization error \ref{fig:10}. When $p_{platy}\approx0.5$, the excess kurtosis is increased in almost all cases, with the average quantization error also enlarged. This supports our claim in Lemma \ref{lemma:hadamard_kurtosis} and Section \ref{sec:kurt_analysis} empirically.

\paragraph{W2A4KV4 Loss Curve Comparison}
\begin{figure}[h!] 
    \centering
    \includegraphics[width=\columnwidth]{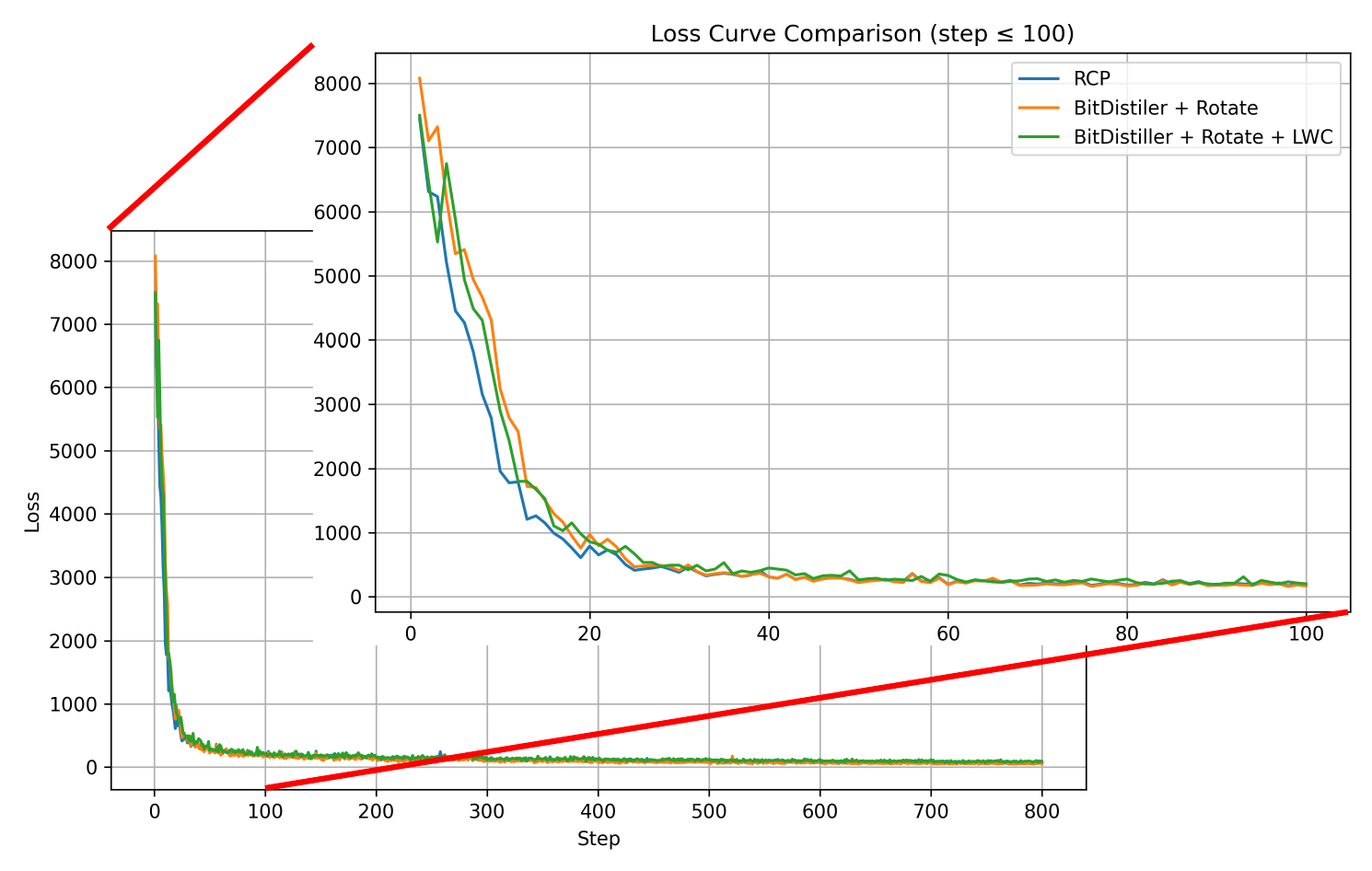} 
    \caption{Comparison of training loss curves for LLaMA-2 7B under W2A4KV4 quantization: BitDistiller + Rotate, BitDistiller + Rotate + LWC, and RCP. }
    \label{fig:loss_curve}
\end{figure}
Figure~\ref{fig:loss_curve} presents the training loss curves of RCP and BitDistiller with rotation, and BitDistiller with rotation and LWC during W2A4KV4 quantization. While all methods eventually converge, a notable performance gap originates from the early stages of training. As highlighted in the magnified region ($\leq 100$ steps), BitDistiller with rotation exhibits pronounced loss spikes, indicating unstable optimization. Even with the addition of LWC, such loss spikes persist, ultimately leading to suboptimal final performance.

In contrast, RCP, which incorporates Learnable Direct Partitioning (LDP), demonstrates significantly more stable loss curve from the beginning of training. These results provide evidence for the necessity of LDP when applying extreme low-bit quantization with rotation and clipping.

\subsection{Additional Experimental Results} 
\label{sec:additional_result}
\paragraph{Implementation Details}
\begin{table}[h]
\centering
\resizebox{1.0\columnwidth}{!}{ 
\renewcommand{\arraystretch}{1.0} 
\setlength{\tabcolsep}{8pt} 
\begin{tabular}{l|c|r|c}
\toprule[1pt]
\textbf{Tensor Type} & \textbf{Sym / Asym} & \textbf{Group Quant.} & \textbf{Clipping Ratio} \\
\hline
Weight & Asym & Yes (Size=128) & LWC \\ 
Activation & Sym  & No (Per-token)     & 0.9  \\
KV-cache   & Asym & Yes (Size=128)     & 0.95 \\
\bottomrule[1pt]
\end{tabular}
}
\caption{Quantization configurations for each component.}
\label{tab:Quant config}
\end{table}

All model parameters are in BF16 format throughout training and evaluation since we observe overflow in the hidden activation of the last two FFNs on several models set to FP16. 

In existing rotation-based PTQ methods~\cite{quarot,spinquant}, rotations are done in FP32 to avoid precision issues. However, this leads to computational overhead due to a large number of typecasting. When fusing rotations to model weights, they are temporarily promoted to FP32, multiplied by an appropriate rotation matrix, and then demoted back to their original precision. For online rotations ($R_2$, $R_3$, and $R_4$), all tensors are processed in BF16.

As shown in Table~\ref{tab:Quant config}, we apply an asymmetric LDP with LWC to weights, a symmetric uniform quantizer to activations, and an asymmetric uniform quantizer with a group size of 128 to the KV-cache, with clipping ratios of 0.9 and 0.95 for activations and KV-cache, respectively.

\paragraph{Additional GEMV Benchmarks}
To compare the gain solely attributed to our non-uniform W2A4 GEMV kernel, we also apply the inefficient quantizer and the online transform to FP16 weights so that the W16A4 model is simulated, and the measured latency values are listed in Table \ref{table:v_proj_latency}. Using online FP16 Hadamard transform, our RCP GEMV is faster than PyTorch \texttt{nn.Linear} kernel, which indicates that our GEMV implementation is faster and can successfully hide its latency to the following activation quantization. 

\subsection{Additional Ablation Studies}
\label{sec:additional_ablation}

\paragraph{Factorized Rotation}\label{sec:factorized_rotation}
\begin{table}[htbp]
\centering
\resizebox{0.9\columnwidth}{!}{ 
\renewcommand{\arraystretch}{1.0} 
\setlength{\tabcolsep}{8pt} 
\begin{tabular}{c|c|c|c|c}
\toprule[1pt]
\textbf{\#Bits} & \textbf{Factorized} & \textbf{Batch} & \textbf{Epoch}&\textbf{PPL$^\downarrow$}\\ \hline
\multirow{2}{*}{W2} &                  & 8      &    8    & 7.6          \\ 
\textbf{}   & \checkmark                 & 1     &    64     & 12.5         \\ \bottomrule[1pt]
\end{tabular}
}
\caption{Comparison of factorized configurations.}
\label{tab:factorized}
\end{table}
In our algorithm, rotation serves as a pre-conditioning tool for reducing outliers in activation and KV-cache. All rotations except the matrices that should be applied online ($R_3$ and $R_4$) are fused into the corresponding model weight at the beginning of the QAT process. This means their orthogonality is not guaranteed during backpropagation steps with AdamW optimizer.

We investigate the impact of preserving the orthogonality of the rotations by modifying the LLaMA-2 model implementation to apply all rotation operators online while freezing the rotation matrices.
Table \ref{tab:factorized} presents the results. Applying factorized rotation prevents the fusion of the rotation matrix into the weight tensor, resulting in an increase in the number of intermediate tensors (rotation matrix and intermediate activation), which significantly raises VRAM requirements. For instance, applying only $R_1$ needs to reduce the training batch size from 8 to 1. Under the condition of maintaining an equal total number of tokens processed by the model, we compared the performance of W2A16KV16 with only $R_1$ applied. The perplexity of BitDistiller with $R_1$ fused was 7.6, whereas applying QAT with factorized rotation resulted in a PPL of 12.5. This indicates that performing weight updates through QAT while preserving $R_1$ orthogonality hinders QAT optimization. This is because the factorization constrains the weight updates to a restricted space defined by the factorized condition, requiring the backpropagation process to maintain within this space. This limitation reduces the flexibility of optimization, making it challenging to efficiently adjust the weights. Consequently, this leads to suboptimal training dynamics and ultimately results in degraded model performance. Furthermore, extending factorization to $R_2$ and $R_4$ would lead to an even greater increase in VRAM usage. In contrast, training fused weight effectively alters only the distribution and is analogous to standard LLM training, which is well-known to perform effectively.
In summary, given that resource consumption increases while performance degrades, we have decided not to explicitly preserve orthogonality and instead allow the algorithm to handle this aspect.
\paragraph{Layerwise vs. End-to-end QAT}
Recent work introduced layerwise QAT~\cite{efficientqat}, which updates one layer at a time while freezing others, allowing training on a single GPU. We extended this approach by applying rotation but observed significant performance degradation. The main issue stemmed from fusing rotation matrices in the weights; layerwise updates disrupted orthogonality, preventing the activation space from restoring its original space, leading to cumulative errors and reduced accuracy. In contrast, end-to-end methods like BitDistiller naturally mitigate this issue during updates. While factorized rotation could help, its high GPU memory requirements for holding rotation matrices and intermediate tensors on GPU memory offsets the advantage. Despite these challenges, exploring single GPU training using rotation matrix remains a promising direction for future work.

\subsection{GEMM Kernel Design for Non-uniform W2A4 Quantization} \label{sec:gemm}
In our initial GEMM implementation, we attempted to leverage the asynchronous copy 
to perform dequantization and MMA operations while loading quantized weights and activations, which resulted in slower performance compared to half-precision PyTorch kernel (approx. 480$\mu$s versus 330$\mu$s on a single (4,096 $\times$ 4,096) linear layer with 2,048 tokens as input). We suggest two underlying reasons; 1) dequantization requires multiple iterations of shifting, masking, and casting to half-precision instruction, and these are typically expensive on the GPU, further deepening the compute-bound nature of the GEMM problem and 2) packing four quantized weights into a single UINT8 and two quantized activation elements into a single INT8 reduces the width of per-block global memory loads, thereby narrowing the chance for latency hiding. Therefore, we decided to leave the prefill acceleration as future work and instead focus on designing a GEMV kernel to accelerate decoding.

\subsection{Details and More Results on GEMV} \label{sec:gemvdetail}

\begin{figure}[h!]
    \centering
    \begin{subfigure}[t]{\linewidth} 
        \centering
        \includegraphics[width=0.9\linewidth]{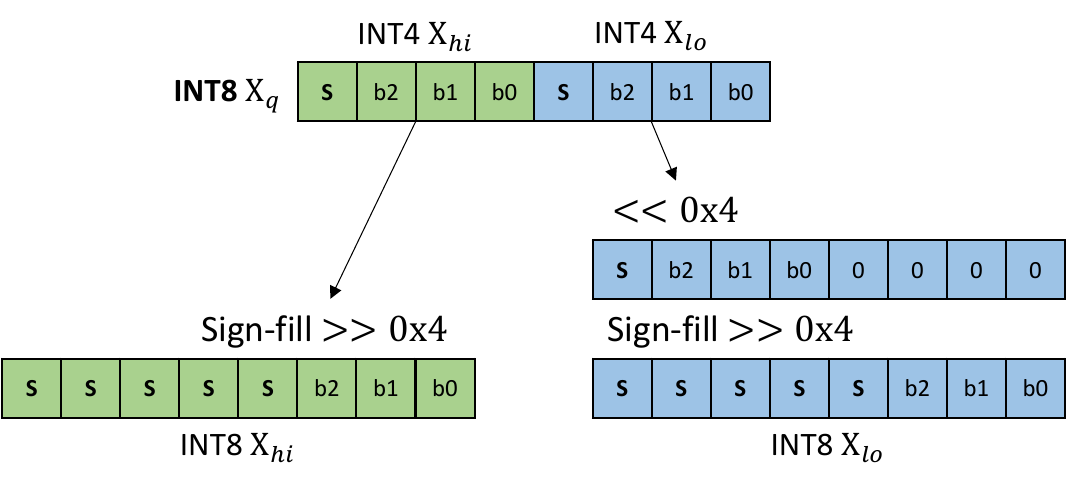} 
        \caption{Dequantization process of two INT4 activations packed in INT8.}
        \label{fig:x_deq}
    \end{subfigure}
    
    \vspace{1em} 
    
    \begin{subfigure}[t]{\linewidth} 
        \centering
        \includegraphics[width=1.0\linewidth]{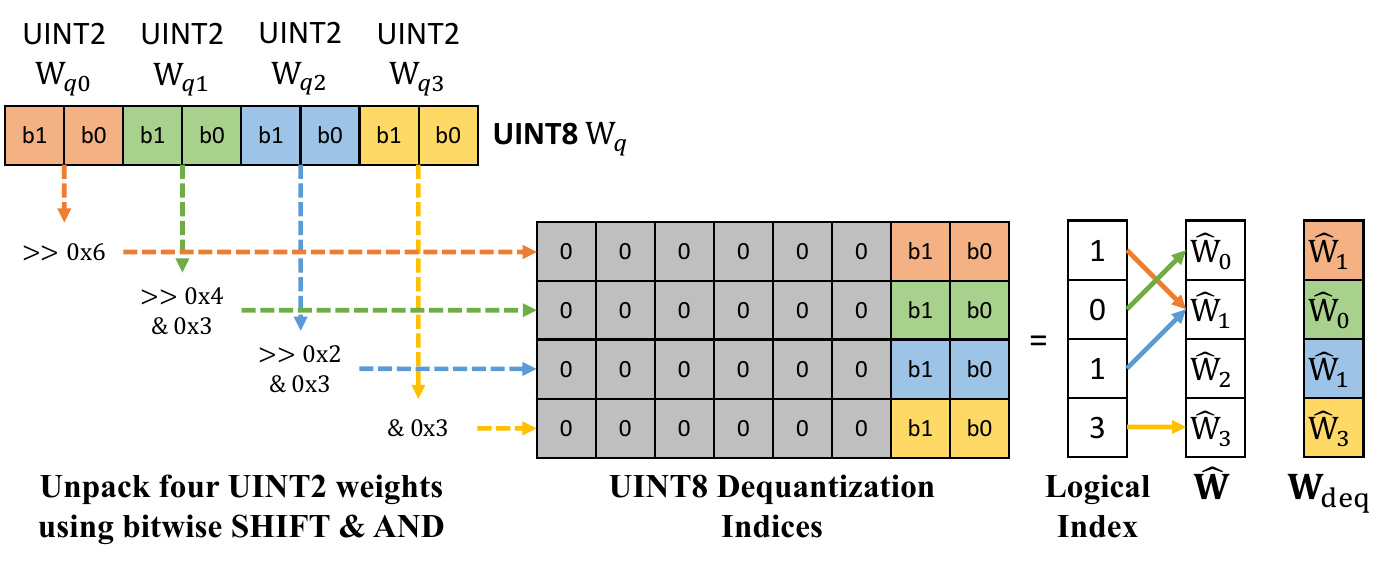} 
        \caption{Dequantization process of 4 UINT2 weights packed in UINT8.}
        \label{fig:w_deq}
    \end{subfigure}
    
    \caption{Online dequantization of INT4 activations and UINT2 weights.}
    \label{fig:deq}
\end{figure}

\paragraph{Block Tiling}
Each thread block consists of 128 threads (4 warps), and we only tile along the output dimension and define the tile size as BH. The reason we do not follow the traditional 2-dimensional tiling is that both the input tokens and weights are stored in row-major format and have sub-byte packing along the column direction, which makes it hard to efficiently use high-bandwidth memory that performs best when reading 128B data consecutively. Also, global loads with small transactions and repeated shared stores complicate the pipeline design for latency hiding and degrade overall performance.

\paragraph{Online Dequantization and Vectorization}
Fig. \ref{fig:deq} illustrates how the activations and weights are dequantized in our GEMV kernel. For activations, there are two INT4 elements ($\text{X}_{hi}, \text{X}_{low}$) in a packed INT8 $\text{X}_q$. For $\text{X}_{hi}$, $\text{X}_q$ is copied to an INT8 register, and the register is right-shifted by 4 bits with sign-filling. For $\text{X}_{low}$, $\text{X}_q$ is also copied to an INT8 register, which is left-shifted by 4 bits first to put the sign bit of $\text{X}_{low}$ to the MSB and then right-shifted by 4 bits with sign filling. This process is shown in Fig. \ref{fig:x_deq}.

For weights, there are four UINT2 elements ($\text{W}_{q0}, \text{W}_{q1}, \text{W}_{q2}, \text{W}_{q3}$) in a packed UINT8 $\text{W}_q$. $\text{W}_q$ is copied to 4 UINT8 registers (for each UINT2 element) that are used as indices to look up the LUT $\hat{\textbf{W}}$. For $\text{W}_{q0}$, the register is right-shifted by 6 bits. For $\text{W}_{q1}$, the register is right-shifted by 4 bits, and a logical AND operation with a bit mask $\texttt{0x03}$ is applied to select only two LSBs. For $\text{W}_{q2}$, the register is right-shifted by 2 bits and also performs logical AND with a bit mask $\texttt{0x03}$. For $\text{W}_{q3}$, the register only does a logical AND with a bit mask $\texttt{0x03}$.

The unit dequantization operations can be vectorized to increase memory throughput so that each thread writes 16B of data to shared memory. For activations, 4 $\text{X}_q$s are loaded from global memory at once by type casting via $\texttt{reinterpret\_cast<char4 *>}$, which produces 8 FP16 dequantized activations to be written in $\text{s}\textbf{X}$. The dequantization is performed the same on each $\text{X}_q$ in a $\texttt{char4}$ struct. For weights, 2 $\text{W}_q$s are loaded from memory via $\texttt{reinterpret\_cast<uint16\_t *>}$. Unlike the activation case, the right-shift and logical AND operation can be naturally iterated 8 times to generate 8 FP16 dequantized weights that are directly multiplied to the corresponding activation from $\text{s}\textbf{X}$.

\paragraph{Shared Epilogue}
As mentioned in Section \ref{sec:gemv_datapath}, a shared output can be necessary due to our chunking strategy. For example, if BH is 4, then two warps will compute one output element to process a weight chunk of size BH/2 × C/4, and after warp-level sum reduction, the reduced values from the two warps must be summed once again. To implement this, we allocate a shared output buffer $\text{s}\textbf{O}$ with twice the number of warps.

After the inner product stage for the first weight chunk, each thread in a block will have an FP32 accumulator with a shape of (4, 32). Applying the warp-shuffle primitive $\texttt{\_\_shfl\_xor\_sync}$ 5 times allows us to sum all accumulations to the first thread of each warp without any global nor shared memory access, producing 4 FP32 values to be cast to FP16 and stored in $\text{s}\textbf{O}[0:4]$. The first and the last two values are summed up as the first and the second output elements, respectively. Repeating the same process on the second weight chunk will produce the next 4 FP32 values for $\text{s}\textbf{O}[4:8]$ to compute the third and the fourth output elements accordingly.

\paragraph{Latency Benchmark}
Our GEMV kernel is fully written in CUDA 12.1 and compiled for Nvidia A100 SXM 40GB model. We build our benchmarking framework upon QuaRot's~\cite{quarot} implementation that provides proper PyTorch bindings and a basic activation quantizer that combines a max reduction function written in PyTorch and a symmetric INT quantizer with INT4 sub-byte data handler from CUTLASS\footnote{https://github.com/NVIDIA/cutlass}.

Since the reduction part is neither a specialized implementation nor compiler-optimized, a huge overhead induced by the QuaRot's activation quantizer is observed (about 100$\mu$s on average). Therefore in the main results, we assume that the symmetric quantization is natively supported by hardware and replace the quantizer with a dummy class that outputs random quantized activation and scale tensors. The results with the inefficient quantizer implementation are listed in Table \ref{tab:gemvQo:v} and \ref{tab:gemvQo:down} for value and down projection weight, respectively. We also report the latency values without activation overhead for the down projection weight in Table \ref{tab:gemvQx:down}.

\begin{table*}[htbp]
    \centering
    \setlength{\tabcolsep}{5pt}
    \renewcommand{\arraystretch}{1.0}
    \resizebox{\textwidth}{!}{%
    {\normalsize
    \begin{tabular}{c cccccccc|c}
        \toprule[2pt]
         \raisebox{-4ex}{\textbf{Model}} & \raisebox{-4ex}{ \textbf{\#Bits \(\text{(W-A-KV)}\)}} & \multicolumn{3}{c}{ \raisebox{-2ex}{\textbf{Configuration}} } & \multicolumn{1}{c}{ \raisebox{-5ex}{\textbf{PIQA}}} & \multicolumn{1}{c}{ \raisebox{-5ex}{\textbf{Hella.}}} & \multicolumn{1}{c}{ \raisebox{-5ex}{\textbf{Wino.}}} & \multicolumn{1}{c}{ \raisebox{-5ex}{\textbf{ARC-c}}} & \multicolumn{1}{c}{ \raisebox{-5ex}{\textbf{Avg.}}}   \\
        \cmidrule(lr){3-5} 
        & & \textbf{Method} & \textbf{Rotation} & \textbf{LDP}  \\
        \midrule
        \multirow{13}{*}{1-7B} & \multirow{1}{*}{16-16-16} &  &  &  & 79.80 & 76.10 & 70.10 & 47.60 & 68.4  \\
        \cmidrule(lr){2-10}
        & \multirow{3}{*}{2-4-16} 
        & BitDistiller &  &  & 61.53 & 35.98 & 49.25 & 23.46 & 43.56  \\
        & & BitDistiller & \checkmark &  & 70.67 & 45.86 & 62.03 & 30.54 & 52.28    \\
        & & RCP & \checkmark & \checkmark & 70.62& 46.41 & 61.48 & 31.32& \textbf{52.46}   \\
        \cmidrule(lr){2-10}
        & \multirow{3}{*}{2-4-4} 
        & BitDistiller &   &   & 63.38 & 34.32 & 50.82 & 23.80 & 43.08   \\
        & & BitDistiller & \checkmark &   & 71.10 & 45.91 & 59.82 & 32.00 & 52.21    \\
        & & RCP & \checkmark & \checkmark & 72.36 & 45.91 & 58.64 & 32.25 & \textbf{52.29}   \\
        \cmidrule(lr){2-10}
        & \multirow{3}{*}{3-4-16} 
        & BitDistiller &   &   & 73.34 & 50.94 & 63.61 & 34.81 & 55.68   \\
        & & BitDistiller & \checkmark &  & 76.71 & 53.96 & 68.19 & 35.23 & 58.52   \\
        & & RCP & \checkmark & \checkmark & 77.20 & 53.11 & 68.43 & 38.82 & \textbf{59.39} \\
        \cmidrule(lr){2-10}
        & \multirow{3}{*}{3-4-4} 
        & BitDistiller &   &   & 73.06 & 50.78 & 65.03 & 35.32 & 56.05    \\
        & & BitDistiller & \checkmark &   & 76.98 & 53.12 & 66.77 & 37.03 & 58.48    \\
        & & RCP & \checkmark & \checkmark & 75.46 & 53.06 & 67.88 & 37.80 & \textbf{58.55}   \\
        \midrule[1.5pt]
        \multirow{13}{*}{2-7B} & \multirow{1}{*}{16-16-16} &  &  &  & 77.86 & 57.14 & 68.35& 43.34 & 61.67 \\
        \cmidrule(lr){2-10}
        & \multirow{3}{*}{2-4-16} 
        & BitDistiller &  &  & 62.95 & 37.33 & 50.20 & 22.95 & 43.36  \\
        & & BitDistiller & \checkmark &  & 70.13 & 45.02 & 60.77 & 30.03 &  \textbf{51.49}   \\
        & & RCP & \checkmark & \checkmark & 69.48 & 45.22 & 59.75 & 29.95 & 51.10 \\
        \cmidrule(lr){2-10}
        & \multirow{3}{*}{2-4-4} 
        & BitDistiller &   &   & 62.70 & 37.18 & 53.91 & 25.93 & 44.93   \\
        & & BitDistiller & \checkmark &  & 69.53 & 45.67 & 59.35 & 29.86 & 51.10   \\
        & & RCP & \checkmark & \checkmark & 69.91 & 44.58 & 59.70 & 30.69 & \textbf{51.22} \\

        \cmidrule(lr){2-10}
        & \multirow{3}{*}{3-4-16} 
        & BitDistiller &   &   & 74.42 & 51.36 & 62.66 & 36.17 & 56.15   \\
        & & BitDistiller & \checkmark &   & 76.06 & 54.26 & 66.45 & 40.35 & 59.28    \\
        & & RCP & \checkmark & \checkmark & 76.65 & 54.25 & 67.80 & 40.35 & \textbf{59.74}   \\
        \cmidrule(lr){2-10}
        & \multirow{3}{*}{3-4-4} 
        & BitDistiller &   &   & 72.41 & 50.51 & 63.29 & 35.83 & 55.51    \\
        & & BitDistiller & \checkmark &   & 76.55 & 53.55 & 65.90 & 39.33 & 58.83    \\
        & & RCP & \checkmark & \checkmark & 76.71 & 53.88 & 65.43 & 41.04 & \textbf{59.27}   \\
        \midrule[1.5pt]
        \multirow{13}{*}{2-13B} & \multirow{1}{*}{16-16-16} &  &  &  & 79.16 & 60.13 & 72.14 & 48.12 & 64.89 \\
        \cmidrule(lr){2-10}
        & \multirow{3}{*}{2-4-16} 
        & BitDistiller &  &  & 61.86 & 33.40 & 53.51 & 23.46 & 43.06 \\
        & & BitDistiller & \checkmark &  & 72.14 & 44.77 & 59.67 & 35.84 & 53.11   \\
        & & RCP & \checkmark & \checkmark & 73.55 & 49.94 & 63.14 & 34.64 & \textbf{55.32}   \\
        \cmidrule(lr){2-10}
        & \multirow{3}{*}{2-4-4} 
        & BitDistiller &   &   & 57.45 & 30.73 & 50.35 & 20.39 & 39.73   \\
        & & BitDistiller & \checkmark &   & 67.68 & 41.58 & 54.62 & 29.69 & 48.39    \\
        & & RCP & \checkmark & \checkmark & 71.65 & 43.79 & 57.30 & 32.68 & \textbf{51.36}   \\
        \cmidrule(lr){2-10}
        & \multirow{3}{*}{3-4-16} 
        & BitDistiller &   &   & 75.29 & 53.91 & 62.50 & 38.56 & 57.57   \\
        & & BitDistiller & \checkmark &  & 77.09 & 56.53 & 70.24 & 44.19 & 62.01   \\
        & & RCP & \checkmark & \checkmark & 77.69 & 57.67 & 70.86 & 45.56 & \textbf{62.95} \\
        \cmidrule(lr){2-10}
        & \multirow{3}{*}{3-4-4} 
        & BitDistiller &   &   & 75.68 & 49.94 & 64.00 & 39.50 & 58.07   \\
        & & BitDistiller & \checkmark &   & 76.71 & 57.11 & 68.03 & 44.19 & \textbf{61.51}    \\
        & & RCP & \checkmark & \checkmark & 77.42 & 56.13 & 69.46 & 42.66 & 61.42   \\
        \bottomrule[2pt]
    \end{tabular}
    }
    }
    \caption{Complete comparison of accuracy on Zero-shot Common Sense Reasoning tasks on LLaMA models. }
\end{table*}

\begin{table*}[htbp]
    \centering
    \setlength{\tabcolsep}{5pt}
    \renewcommand{\arraystretch}{1.0}
    \resizebox{\textwidth}{!}{%
    {\normalsize
    \begin{tabular}{c cccccccc|c}
        \toprule[2pt]
         \raisebox{-4ex}{\textbf{Model}} & \raisebox{-4ex}{ \textbf{\#Bits \(\text{(W-A-KV)}\)}} & \multicolumn{3}{c}{ \raisebox{-2ex}{\textbf{Configuration}} } & \multicolumn{1}{c}{ \raisebox{-5ex}{\textbf{PIQA}}} & \multicolumn{1}{c}{ \raisebox{-5ex}{\textbf{Hella.}}} & \multicolumn{1}{c}{ \raisebox{-5ex}{\textbf{Wino.}}} & \multicolumn{1}{c}{ \raisebox{-5ex}{\textbf{ARC-c}}} & \multicolumn{1}{c}{ \raisebox{-5ex}{\textbf{Avg.}}}   \\
        \cmidrule(lr){3-5} 
        & & \textbf{Method} & \textbf{Rotation} & \textbf{LDP}  \\
        \midrule
        \multirow{13}{*}{3.2-1B} & \multirow{1}{*}{16-16-16} &  &  &  & 75.30 & 60.70 & 60.90 & 38.70 & 58.90  \\
        \cmidrule(lr){2-10}
        & \multirow{3}{*}{2-4-16} 
        & BitDistiller &  &  & 51.95 & 27.41 & 48.46 & 19.45 & 36.82  \\
        & & BitDistiller & \checkmark &  & 61.15 & 30.66 & 50.67 & 21.84 &  41.08   \\
        & & RCP & \checkmark & \checkmark & 61.42& 31.55 & 51.78 & 20.65& \textbf{41.08}   \\
        \cmidrule(lr){2-10}
        & \multirow{3}{*}{2-4-4} 
        & BitDistiller &   &   & 55.33 & 26.62 & 48.46 & 19.79 & 37.55   \\
        & & BitDistiller & \checkmark &   & 61.75 & 30.05 & 51.22 & 20.05 & 40.77    \\
        & & RCP & \checkmark & \checkmark & 60.71 & 31.54 & 53.51 & 21.42 & \textbf{41.80}   \\
        \cmidrule(lr){2-10}
        & \multirow{3}{*}{3-4-16} 
        & BitDistiller &   &   & 53.53 & 28.35 & 48.61 & 19.62 & 37.53   \\
        & & BitDistiller & \checkmark &  & 69.53 & 40.31 & 55.40 & 26.27 & 47.88   \\
        & & RCP & \checkmark & \checkmark & 69.64 & 40.57 & 56.12 & 26.37 & \textbf{48.18} \\
        \cmidrule(lr){2-10}
        & \multirow{3}{*}{3-4-4} 
        & BitDistiller &   &   & 54.18 & 28.26 & 50.90 & 21.67 & 38.75    \\
        & & BitDistiller & \checkmark &   & 68.98 & 37.80 & 55.40 & 26.36 & 47.14    \\
        & & RCP & \checkmark & \checkmark & 68.12 & 39.30 & 56.12 & 26.11 & \textbf{47.41}   \\
        \midrule[1.5pt]
        \multirow{13}{*}{3.2-3B} & \multirow{1}{*}{16-16-16} &  &  &  & 76.00 & 71.00 & 66.60 & 47.60 & 65.30 \\
        \cmidrule(lr){2-10}
        & \multirow{3}{*}{2-4-16} 
        & BitDistiller &  &  & 54.02 & 26.80 & 52.48 & 18.25 & 37.89  \\
        & & BitDistiller & \checkmark &  & 65.99 & 36.51 & 52.48 & 26.19 & 45.29  \\
        & & RCP & \checkmark & \checkmark & 65.43 & 37.35 & 54.70 & 25.43 & \textbf{45.71} \\
        \cmidrule(lr){2-10}
        & \multirow{3}{*}{2-4-4} 
        & BitDistiller &   &   & 51.84 & 26.70 & 51.38 & 19.11 & 37.26   \\
        & & BitDistiller & \checkmark &  & 64.30 & 36.26 & 51.38 & 25.08 & 44.26   \\
        & & RCP & \checkmark & \checkmark & 65.45 & 36.66 & 53.75 & 26.37 & \textbf{45.56} \\

        \cmidrule(lr){2-10}
        & \multirow{3}{*}{3-4-16} 
        & BitDistiller &   &   & 52.72 & 26.66 & 50.43 & 19.45 & 37.32   \\
        & & BitDistiller & \checkmark &   & 74.04 & 49.56 & 63.22 & 35.83 & 55.66    \\
        & & RCP & \checkmark & \checkmark & 73.77 & 49.52 & 62.65 & 37.54 & \textbf{55.87}   \\
        \cmidrule(lr){2-10}
        & \multirow{3}{*}{3-4-4} 
        & BitDistiller &   &   & 53.91 & 26.82 & 48.03 & 20.30 & 37.27    \\
        & & BitDistiller & \checkmark &   & 74.31 & 49.19 & 60.06 & 36.77 & 55.08    \\
        & & RCP & \checkmark & \checkmark & 73.18 & 48.87 & 62.43 & 36.01 & \textbf{55.12}   \\
        \midrule[1.5pt]
        \multirow{13}{*}{3-8B} & \multirow{1}{*}{16-16-16} &  &  &  & 80.70 & 79.60 & 73.70 & 57.70 & 72.93 \\
        \cmidrule(lr){2-10}
        & \multirow{3}{*}{2-4-16} 
        & BitDistiller &  &  & 57.23 & 29.96 & 49.48 & 21.16 & 39.46 \\
        & & BitDistiller & \checkmark &  & 69.96 & 44.30 & 59.43 & 28.66 & 50.59   \\
        & & RCP & \checkmark & \checkmark & 69.16 & 44.67 & 59.91 & 29.69 & \textbf{50.86}   \\
        \cmidrule(lr){2-10}
        & \multirow{3}{*}{2-4-4} 
        & BitDistiller &   &   & 56.42 & 29.57 & 52.09 & 20.90 & 39.75   \\
        & & BitDistiller & \checkmark &   & 69.15 & 43.62 & 57.85 & 28.58 & 49.80    \\
        & & RCP & \checkmark & \checkmark & 69.97 & 44.32 & 59.51 & 27.82 & \textbf{50.41}   \\
        \cmidrule(lr){2-10}
        & \multirow{3}{*}{3-4-16} 
        & BitDistiller &   &   & 72.47 & 49.72 & 62.43 & 36.94 & 55.39   \\
        & & BitDistiller & \checkmark &  & 77.25 & 55.18 & 68.90 & 42.91 & 61.06   \\
        & & RCP & \checkmark & \checkmark & 77.64 & 55.21 & 69.93 & 43.34 & \textbf{61.53} \\
        \cmidrule(lr){2-10}
        & \multirow{3}{*}{3-4-4} 
        & BitDistiller &   &   & 73.32 & 49.97 & 64.87 & 37.45 & 56.35   \\
        & & BitDistiller & \checkmark &   & 75.35 & 53.95 & 67.64 & 41.80 & 59.69    \\
        & & RCP & \checkmark & \checkmark & 76.16 & 54.35 & 71.19 & 42.75 & \textbf{61.11}   \\
        \bottomrule[2pt]
    \end{tabular}
    }
    }
    \caption{Complete comparison of accuracy on Zero-shot Common Sense Reasoning tasks on LLaMA models. }
\end{table*}

\begin{table*}[htbp]
\resizebox{\textwidth}{!}{%
\begin{tabular}{c|c|c|c|c|c|c|c|c}
\toprule[2pt]
\textbf{} & \textbf{hotpotqa} & \textbf{mqa\_en} & \textbf{triviaqa} & \textbf{2wikimqa} & \textbf{musique} & \textbf{samsum} & \textbf{passage\_count} & \textbf{Avg.} \\ \hline
\textbf{FP16} & 30.45 & 33.76 & 85.72 & 26.32 &9.74 & 37.74 & 4.0 & 32.53  \\ \hline
\textbf{BitDistiller} & 2.95 & 11.09 & 7.28 & 4.42 & 2.03 & 1.97 & 0.86 & 4.37 \\ \hline
\textbf{RCP} &10.42 & 26.73 & 41.77 & 17.48 &4.12 & 33.31 & 1.43 & \textbf{19.32} \\ 
\bottomrule[2pt]
\end{tabular}
}
\caption{Performance comparison on the LongBench dataset. W2A4KV4 quantization is applied to the \textbf{LLaMA-2-7B-chat-4k} model. }
\label{tab:4k-long}
\end{table*}

\begin{table*}[htbp]
\resizebox{\textwidth}{!}{%
\begin{tabular}{c|c|c|c|c|c|c|c|c}
\toprule[2pt]
\textbf{} & \textbf{hotpotqa} & \textbf{mqa\_en} & \textbf{triviaqa} & \textbf{2wikimqa} & \textbf{musique} & \textbf{samsum} & \textbf{passage\_count} & \textbf{Avg.} \\ \hline
\textbf{FP16} & 15.74 & 24.07 & 84.67 & 13.8 & 8.81 & 42.73 & 0.07 & 27.13  \\ \hline
\textbf{BitDisitller} & 2.18 & 9.72 & 11.82 & 5.18 & 1.09 & 5.65 & 0.48 & 5.16 \\ \hline
\textbf{RCP} & 5.97 & 13.58 & 33.07 & 10.81 & 2.1 & 19.34 & 1.13 & \textbf{12.29} \\ 
\bottomrule[2pt]
\end{tabular}
}

\caption{Performance comparison on the LongBench dataset. W2A4KV4 quantization is applied to the \textbf{LLaMA-2-7B-Instruct-32k} model. }
\label{tab:32k-long}
\end{table*}

\begin{table*}[htbp]
\resizebox{\textwidth}{!}{%
\centering
\renewcommand{\arraystretch}{1} 
\begin{tabular}{c|cccccc}
\toprule[2pt]
\textbf{Layer Size}  & \textbf{W16A4} & \textbf{W16A4+FP32Had}&\textbf{W16A4+FP16Had} & \textbf{RCP}&\textbf{RCP+FP32Had} &\textbf{RCP+FP16Had} \\ \hline
(2048, 2048) &  0.168 & 0.274 & 0.248 & 0.131 & 0.248 & 0.214  \\
(2048, 8192) &  0.327 & 0.387 & 0.348 & 0.143 & 0.240 & 0.218  \\
(3072, 3072) &  0.228 & 0.483 & 0.373 & 0.131 & 0.295 & 0.265  \\
(3072, 8192) &  0.526 & 0.773 & 0.661 & 0.140 & 0.294 & 0.271  \\
(4096, 4096) &  0.369 & 0.510 & 0.398 & 0.133 & 0.250 & 0.221  \\
(4096, 11008) &  0.866 & 1.014 & 0.902 & 0.143 & 0.250 & 0.223  \\
(4096, 14336) &  1.108 & 1.255 & 1.146 & 0.142 & 0.247 & 0.226  \\
\bottomrule[2pt]
\end{tabular}
}
\caption{GEMV latency for the value projection is measured with the overhead from activation quantization. The layer size is composed as (input channel, output channel). All latency numbers are in milliseconds.}
\label{table:v_proj_latency}
\end{table*}
\begin{table*}[htbp]
\resizebox{\textwidth}{!}{%
\centering
\renewcommand{\arraystretch}{1} 
\begin{tabular}{c|ccccccc}
\toprule[2pt]
\textbf{Layer Size} & \textbf{FP16} & \textbf{RCP} & \textbf{RCP+FP16Had}&\textbf{RCP+FP32Had} & \textbf{QuaRot}&\textbf{QuaRot+FP16Had} &\textbf{QuaRot+FP32Had} \\ \hline
(2048, 8192)  & 0.054 &  0.036 & 0.110 & 0.146 & 0.073 & 0.155 & 0.186  \\
(3072, 8192)  & 0.054 & 0.035 & 0.169 & 0.198 & 0.074 & 0.212 & 0.237  \\
(4096, 11008) & 0.077 &  0.048 & 0.120 & 0.148 & 0.088 & 0.157 & 0.186  \\
(4096, 14336) & 0.110 &  0.059 & 0.121 & 0.149 & 0.079 & 0.157 & 0.183  \\
\bottomrule[2pt]
\end{tabular}
}
\caption{GEMV latency for the down projection is measured except activation quantization overhead. The layer size is composed as (input channel, output channel). All latency numbers are in milliseconds.}
\label{tab:gemvQx:down}
\end{table*}

\begin{table*}[htbp]
\resizebox{\textwidth}{!}{%
\centering
\renewcommand{\arraystretch}{1} 
\begin{tabular}{c|cccccc}
\toprule[2pt]
\textbf{Layer Size}  & \textbf{RCP} & \textbf{RCP+FP16Had}&\textbf{RCP+FP32Had} & \textbf{QuaRot}&\textbf{QuaRot+FP16Had} &\textbf{QuaRot+FP32Had} \\ \hline
(2048, 2048) &  0.131 & 0.214 & 0.248 & 0.170 & 0.248 & 0.276  \\
(3072, 3072) &  0.131 & 0.265 & 0.295 & 0.168 & 0.304 & 0.331  \\
(4096, 4096) &  0.133 & 0.221 & 0.250 & 0.174 & 0.250 & 0.282  \\
\bottomrule[2pt]
\end{tabular}
}
\caption{GEMV latency for the value projection is measured including activation quantization overhead. The layer size is composed as (input channel, output channel). All latency numbers are in milliseconds.}
\label{tab:gemvQo:v}
\end{table*}

\begin{table*}[htbp]
\resizebox{\textwidth}{!}{%
\centering
\renewcommand{\arraystretch}{1} 
\begin{tabular}{c|cccccc}
\toprule[2pt]
\textbf{Layer Size}  & \textbf{RCP} & \textbf{RCP+FP16Had}&\textbf{RCP+FP32Had} & \textbf{QuaRot}&\textbf{QuaRot+FP16Had} &\textbf{QuaRot+FP32Had} \\ \hline
(2048, 8192) &  0.143 & 0.218 & 0.240 & 0.186 & 0.261 & 0.289  \\
(3072, 8192) &  0.140 & 0.271 & 0.294 & 0.177 & 0.318 & 0.340  \\
(4096, 11008) &  0.143 & 0.223 & 0.250 & 0.177 & 0.264 & 0.288  \\
(4096, 14336) &  0.142 & 0.226 & 0.247 & 0.177 & 0.259 & 0.285  \\
\bottomrule[2pt]
\end{tabular}
}
\caption{GEMV latency for the down projection is measured including activation quantization overhead. The layer size is composed as (input channel, output channel). All latency numbers are in milliseconds.}
\label{tab:gemvQo:down}
\end{table*}

\subsection{Reasoning Task Example: HumanEval}
\begin{figure*}[] 
    \centering
    \includegraphics[width=\textwidth]{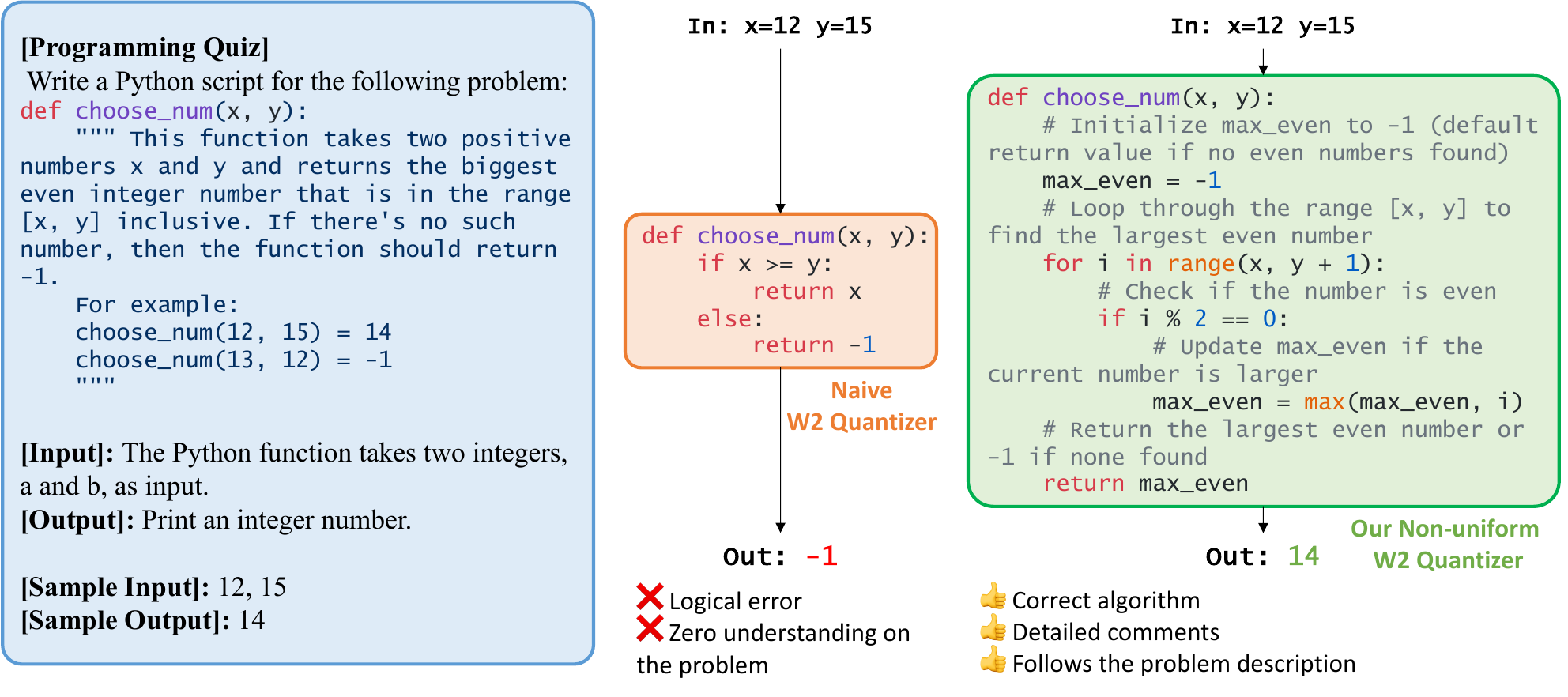} 
    \caption{A reasoning task example from HumanEval ~\cite{humaneval} benchmark, conducted by two differently quantized WizardCoder 7B ~\cite{luo2023wizardcoder} models. The results in the orange box is from state-of-the-art QAT method BitDistiller~\cite{bitdistiller} with applying rotation. In the green box, our proposed RCP is applied. Both methods employ exactly the same 4-bit quantization setting for activation and KV-cache.}
    \label{fig:coding}
\end{figure*}

We evaluate the capability of the WizardCoder 7B model to generate solutions for coding problems. The results are presented in Fig. \ref{fig:coding}. The orange box in Fig. \ref{fig:coding} represent the model output after applying rotation and quantizing the weights to W2A4KV4 using a uniform asymmetric quantizer. Under uniform quantization, it is evident that the model fails to perform logical generation tasks even applying rotation; it merely produces the structural template of code without generating functionality correct code. In contrast, the green box shows the results when the weights are quantized to W2A4KV4 using LDP. Unlike the uniform quantizer, the LDP approach yields code that not only adheres faithfully to the given instructions and generates a functionality correct algorithm, but also provides detailed explanatory comments. While perplexity on standard language modeling tasks did not reveal significant differences between the two cases, these findings suggest that LDP plays a crucial role in enabling logical reasoning tasks under extreme low-bit quantization.

\begin{figure*}[] 
    \centering
    \includegraphics[width=\textwidth]{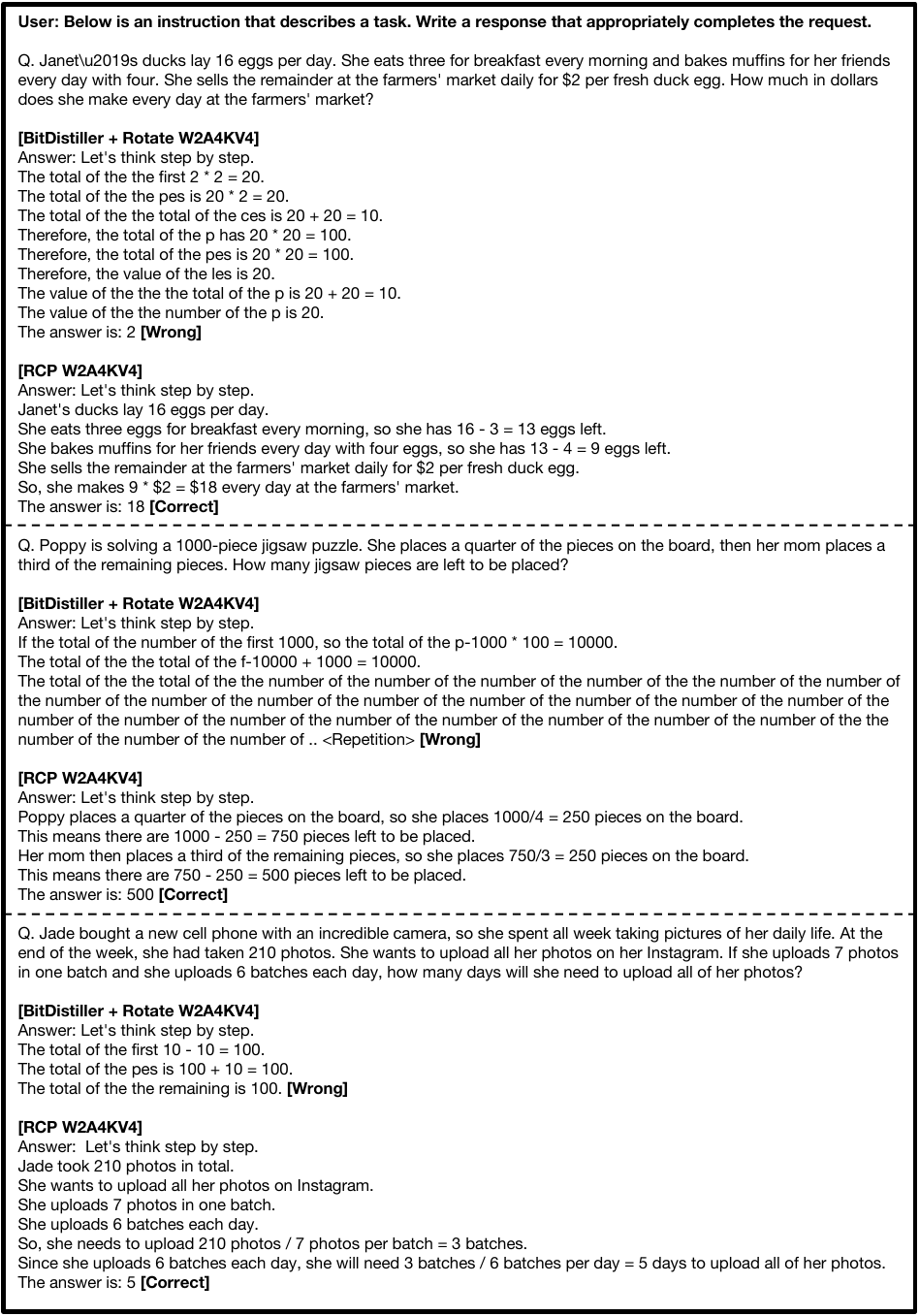} 
    \caption{Comparison of reasoning ouptut on GSM8K using MetaMath 7B under W2A4KV4.}
    \label{fig:gsm8k_full}
\end{figure*}
\subsection{Information About Use of AI Assistants}
AI assistance was strictly limited to linguistic perspectives, such as grammar and spell checking, and finding synonyms.
\end{document}